\pdfoutput=1
\documentclass{article}

\usepackage[preprint]{neurips_2024}
\usepackage[utf8]{inputenc} 
\usepackage[T1]{fontenc}    
\usepackage[hidelinks]{hyperref}
\usepackage{url}            
\usepackage{booktabs}       
\usepackage{amsfonts}       
\usepackage{nicefrac}       
\usepackage{microtype}      
\usepackage{xcolor}         
\usepackage{bbm}

\usepackage[algo2e]{algorithm2e}
\usepackage{textcomp}
\usepackage{algorithm}
\usepackage{algpseudocode}
\usepackage{graphicx}
\graphicspath{ {./images/} }
\usepackage{subcaption}
\usepackage{cleveref}
\usepackage[title]{appendix}

\setcitestyle{numbers,square}

\title{METRIK: Measurement-Efficient Randomized Controlled Trials using Transformers with Input Masking}

\author{
  Sayeri Lala$^{1}$ \qquad Niraj K. Jha$^{1}$ \\
  $^1$Department of Electrical and Computer Engineering \\
  Princeton University\\
  Princeton, NJ 08540 \\
  \texttt{\{slala,jha\}@princeton.edu}
}

\begin{document}

\maketitle

\begin{abstract}
Clinical randomized controlled trials (RCTs) collect hundreds of measurements spanning various metric types 
(e.g., laboratory tests, cognitive/motor assessments, etc.) across 100s-1000s of subjects to 
evaluate the effect of a treatment, but do so at the cost of significant trial expense. To reduce the number of measurements, trial protocols 
can be revised to remove metrics extraneous to the study's objective, but doing so requires additional human labor 
and limits the set of hypotheses that can be studied with the collected data. In contrast, a planned missing design 
(PMD) can reduce the amount of data collected without removing any metric by imputing the unsampled data. Standard 
PMDs randomly sample data to leverage statistical properties of imputation algorithms, but are \textit{ad hoc}, hence suboptimal. Methods that 
learn PMDs produce more sample-efficient PMDs, but are not suitable for RCTs because they require ample prior 
data (150+ subjects) to model the data distribution. Therefore, we introduce a framework called 
\underline{M}easurement \underline{E}fficien\underline{T} \underline{R}andomized Controlled Trials using 
Transformers with \underline{I}nput Mas\underline{K}ing (METRIK), which, for the first time, calculates a PMD 
specific to the RCT from a modest amount of prior data (e.g., 60 subjects). Specifically, METRIK models the PMD as a 
learnable input masking layer that is optimized with a state-of-the-art imputer based on the Transformer 
architecture. METRIK implements a novel sampling and selection algorithm to generate a PMD that satisfies the 
trial designer's objective, i.e., whether to maximize sampling efficiency or imputation performance for a given 
sampling budget. Evaluated across five real-world clinical RCT datasets, METRIK increases the sampling efficiency of and imputation performance under the generated PMD by leveraging correlations over time and across metrics, thereby removing the need to manually remove metrics from the RCT. \end{abstract}

\section{Introduction} \label{sec:intro}
The randomized controlled trial (RCT) is the gold-standard approach for characterizing treatment effects of new medical interventions (e.g., drugs) and is commonly used by Phase-3 clinical trials conducted for market approval \citep{friedman2015fundamentals}. However, RCTs are expensive (e.g., Phase-3 trials cost USD 12M-53M across different therapeutic areas over years 2004-2012), with clinical procedures/laboratory tests being a major cost driver, accounting for at least 25\% of trial expenses \citep{sertkaya2016key}. This is not surprising, given that RCTs monitor numerous metrics spanning various categories, such as demographics, quality of life, and medical history \citep{friedman2015fundamentals}, which translate into at least hundreds of measurements collected per subject over the course of a trial \citep{getz2013drug}. While internal governance committees have been established to simplify data collection protocols by removing extraneous metrics \citep{getz2013new}, this approach requires additional time and labor and is not ideal since extraneous metrics are often useful for generating new hypotheses \citep{friedman2015fundamentals}. 

Instead of removing metrics, we propose implementing a planned missing design (PMD) that removes measurements across metrics over the course of the study. While methods for constructing a PMD have been developed and applied to epidemiological studies, their application to RCTs has been limited because existing PMD strategies either require human input or ample prior data (e.g., from 150+ subjects) \citep{rioux2020reflection}, making these methods unsuitable for RCTs that examine new interventions. In this study, we present a new framework called 
\underline{M}easurement \underline{E}fficien\underline{T} \underline{R}andomized Controlled Trials using 
Transformers with \underline{I}nput Mas\underline{K}ing (METRIK), which learns a PMD 
specific to the RCT using only data obtained through a small pilot study (e.g., with 60 subjects). In contrast to existing approaches that learn PMDs using models fit on large prior datasets \citep{adiguzel2008split,wu2016search,imbriano2018methods}, METRIK learns the PMD from a small dataset by using a performant imputation architecture \citep{zerveas2021transformer}, based on the Transformer \citep{vaswani2017attention}, combined with a differentiable mask layer \cite{csordas2021neural} that represents the PMD, which are trained with a version of the masked autoencoding objective adapted for time-series data \citep{zerveas2021transformer}. By sampling over hyperparameters that influence the properties of the learned PMD, METRIK generates a diverse pool of candidate PMDs, which it then mines to identify PMDs that satisfy the designer's objective of either maximizing the PMD's sampling efficiency or imputation performance at some efficiency level.

We demonstrate the effectiveness of METRIK across several real-world RCT datasets and show its advantages over existing \textit{ad hoc} PMD strategies based on random sampling. We provide additional analyses in the form of ablation studies and samples of generated PMDs to shed light on how METRIK improves efficiency/imputation performance.


The rest of the paper is organized as follows. Sec.~\ref{sec:rel_works} discusses related works. Sec.~\ref{sec:method} presents the METRIK framework. Sec.~\ref{sec:exp_setup} presents the experimental setup. Sec.~\ref{sec:results} presents and discusses the experimental results. Sec.~\ref{sec:limitations} discusses limitations of the framework. Sec.~\ref{sec:conclusion} draws conclusions. 

\section{Related Works} \label{sec:rel_works}

In this section, we review works related to PMD generation and imputation. 

\subsection{Strategies for PMD Generation}
Methods for constructing PMDs fall under random sampling designs (RSDs) and optimization-based approaches \citep{rioux2020reflection}. RSDs leverage statistical properties of imputation algorithms to implement PMDs that yield unbiased parameter estimates \cite{rioux2020reflection}. While the matrix sampling design randomly samples each measurement per subject \citep{raghunathan1995split}, tailored randomized PMDs can further reduce the variance of certain types of parameter estimates. For example, the Multiform (MF) design \citep{graham2006planned} yields more efficient estimates of correlations by dividing measurements or questions into sets, creating multiple forms that pool questions across random pairs of sets, and randomly administering forms across subjects. However, because the MF design reduces the variance of the estimates unequally across the pairs of variables, it requires users to allocate variables across sets to ensure that all interactions are measured efficiently. While \textit{ad hoc} allocation strategies have been suggested, some studies have shown that the allocation can be optimized \citep{adiguzel2008split, wu2016search, imbriano2018methods}, but these approaches require fitting a data-generating model, which requires large prior datasets (e.g., 500 subjects), making these methods unsuitable for RCTs.   

\subsection{Imputation Algorithms}

Basic imputation techniques, e.g., mean filling or univariate approaches, are unable to model complex relationships over variables in multivariate datasets \citep{van2018flexible}. Hence, imputation algorithms tailored to the multivariate setting have been developed. For example, the MICE algorithm uses a Markov Chain Monte Carlo approach to learn the joint distribution over variables, thereby removing the need to pre-specify the form of the distribution that is needed with joint modeling \citep{van2011mice} . Specifically, at each iteration, each missing variable is regressed using models (e.g., linear regressors) fit on the observed and imputed variables from a previous iteration. 3D-MICE \citep{luo20183d} extends MICE to integrate temporal information present over the variables as in clinical data collection settings. Specifically, it uses Gaussian processes to model temporal correlations and combines predictions under the Gaussian process model with predictions under MICE using a weighted average, where weights are determined by the variance of the predictions. Later work has shown that machine-learning-based approaches, e.g., gradient boosting, outperform 3D-MICE on clinical imputation tasks, but these methods use hand-designed features. In contrast, approaches based on deep learning architectures, e.g., a recurrent neural network (RNN), can perform competitively to machine-learning approaches \citep{luo2022evaluating} and demonstrate state-of-the-art performance on various clinical imputation tasks \citep{liu2023handling} without requiring features to be pre-specified. Specifically, one study demonstrates that a Transformer-based architecture developed for multivariate time-series applications (MVTS) \citep{zerveas2021transformer} outperforms RNN-based imputation frameworks on a clinical imputation task \citep{yildiz2022multivariate}, given the Transformer's abilities to model long-range dependencies \citep{vaswani2017attention} and the use of masked autoencoding to pre-train the entire model \citep{zerveas2021transformer}.  

\begin{figure}[H]
    \centering
    \begin{subfigure}[b]{\textwidth}
        \centering
        \includegraphics[width=\textwidth]{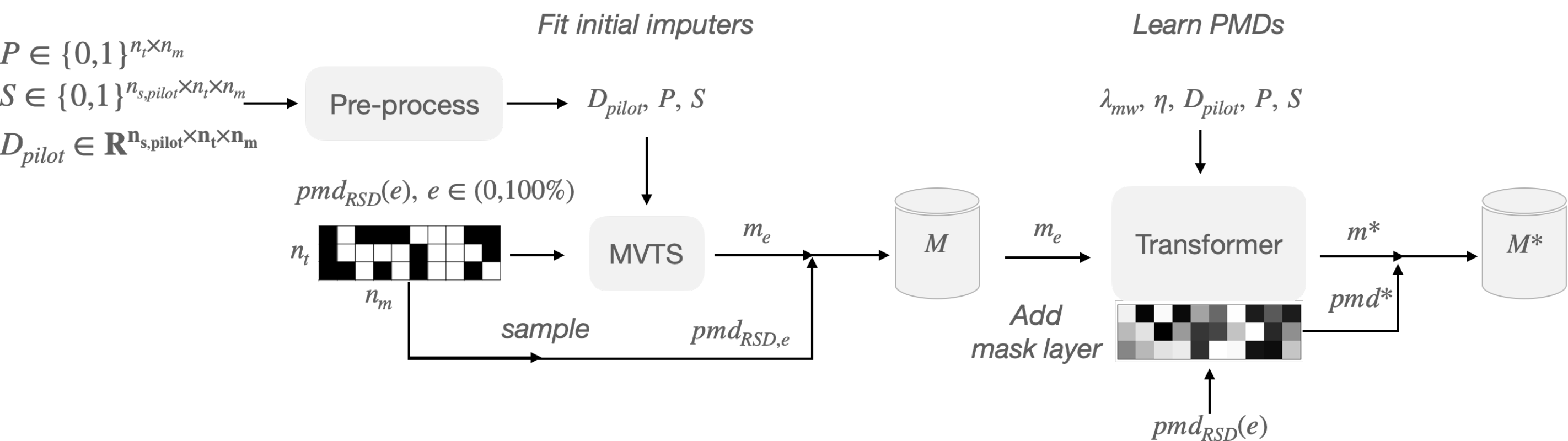}
        \caption{Flowchart for generating candidate PMDs.}
         \label{fig:gen_cand}
    \end{subfigure}
    
    \begin{subfigure}[b]{\textwidth}
        \centering
        \includegraphics[width=\textwidth,angle=270,scale=0.2]{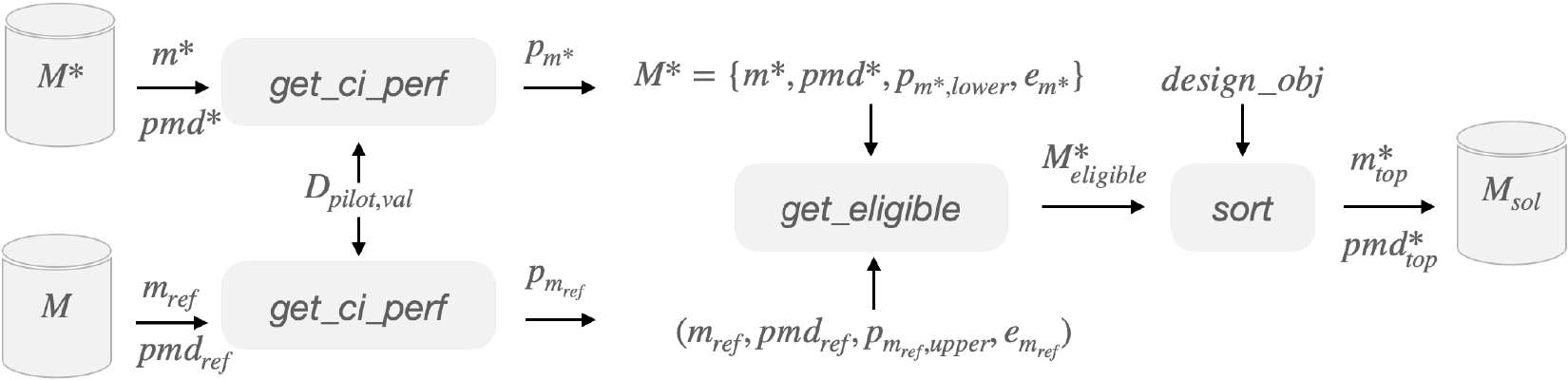}
        \caption{Flowchart for choosing PMDs.}
         \label{fig:choose_pmds}
    \end{subfigure}
    \caption{Flowchart of the METRIK framework.}
    \label{fig:flowchart}
\end{figure}

\section{The METRIK Framework} \label{sec:method}

METRIK is a framework for learning a PMD for an RCT from a small internal pilot study \citep{friedman2015fundamentals}. Its flowchart is shown in Fig.~\ref{fig:flowchart}. METRIK generates candidate PMDs by first pre-processing the pilot study data to make it suitable for training, fitting initial imputation models using the MVTS algorithm, and then training these models further with mask weight learning to generate a diverse set of PMDs, as shown in Fig.~\ref{fig:gen_cand}. Then, it chooses PMDs and their associated imputers among the candidates by ranking them according to the user's design objective, i.e., whether to maximize efficiency or imputation performance, as shown in Fig.~\ref{fig:choose_pmds}.  

Next, we delve into details underlying each step of the framework.

\subsection{Data Collection and Pre-processing}

METRIK requires that the trial investigator implement an internal pilot study \citep{friedman2015fundamentals}, which collects data using a relatively small sample size (e.g., 60 \citep{teare2014sample} vs. a typical RCT size of 100-1000 patients \cite{stanley2007design}) using the same data collection protocol as the planned RCT. After data collection, METRIK pre-processes the pilot study dataset, $D_{pilot}$, by handling native missingness, normalizing it, and dividing it into training and validation sets. Native missingness stems from patient dropout \citep{friedman2015fundamentals} and sampling irregularities in the data collection protocol across metrics.  METRIK addresses it through mean/mode imputation. The missingness is determined by $P \in \{0,1\}^{n_t \times n_m}$, a binary mask indicating which measurements from the data collection protocol are eligible for masking, where $n_t$ is the study duration and $n_m$ is the number of metrics monitored over the study (this mask is provided by the trial designer), and $S \in \{0,1\}^{n_{s,pilot} \times n_t \times n_m}$, a binary mask indicating which measurements were collected per subject, where $n_{s,pilot}$ is the number of subjects monitored over the pilot study (this mask is also provided by the trial designer). After imputation, it performs min-max normalization per metric and then divides the dataset into training and validation sets at the patient level.   

\subsection{Generating PMDs}

Next, we walk through the flowchart for generating candidate PMDs, as shown in Fig.~\ref{fig:gen_cand} (pseudocode given in Appendix~\ref{app:pseudocode}, Alg.~\ref{alg:gen_cand}). To generate candidate PMDs, METRIK seeds PMD learning by fitting initial imputers on the pilot study dataset. Specifically, METRIK trains an imputer $m_e$ for a given efficiency level $e$, defined as the fraction of measurements collected under study protocol $P$ that are masked out by the PMD. $m_e$ is trained using the MVTS algorithm, given its state-of-the-art performance on time-series-based applications. Specifically, it uses an RSD-based PMD generator, given by $pmd_{RSD}(e)$, which generates random binary matrices of shape $n_t \times n_m$ using a Bernoulli distribution parameterized by efficiency level $e$ while being constrained by $P$. METRIK trains imputers across masking efficiencies $e$ ranging between 0 and 100\%. During training, losses are only evaluated over eligible masked elements, determined by $P$ and $S$. METRIK stores the trained models along with samples of corresponding RSD-based PMDs in set $M$.  

METRIK then uses the initial imputation models to learn a new set of models and model-optimized PMDs. METRIK achieves this using the mask learning algorithm \citep{csordas2021neural}, which learns a binary mask by modeling it as a differentiable layer of logits and fitting it with the task objective along with regularization applied to the logits (mask weights). METRIK applies the mask learning algorithm to each initial imputer and generates a set of diverse PMDs per imputation model $m_e$ by sampling different hyperparameter combinations of mask regularization weights and learning rates used to train the mask-modulated model. Specifically, for each sampled mask regularization weight $\lambda_{mw}$ and learning rate $\eta$, METRIK attaches a mask layer \cite{csordas2021neural} to the bottom of each initial imputer, where the same Transformer architecture is used for the imputer with weights initialized from $m_e$. The weight mask is initialized according to the distribution used to generate an RSD-based PMD of efficiency $e$ in order to mitigate potential effects from distribution shifts during training. The masked imputer is then trained on the imputation task induced by the parameterized PMD using the MVTS algorithm, pilot study dataset, and sampled hyperparameters. In contrast to the original work \citep{csordas2021neural}, we do not freeze imputer weights since our goal is to optimize performance. In addition, during training, no mask noise is added and gradients are zeroed out for measurements excluded from the data protocol to ensure that the derived PMD satisfies constraints set by $P$. After training, the learned PMD $pmd^*$ (i.e., represented by the mask layer) and imputer $m^*$ get stored in set $M^*$. 


\subsection{Choosing PMDs} \label{sec:choose_pmds}

Given the sets of initial and learned imputation models and their associated PMDs, i.e., $M$ and $M^{*}$, respectively, METRIK chooses imputer-PMD pairs that are optimal based on the design criterion, as shown in the flowchart in Fig.~\ref{fig:choose_pmds} (pseudocode given in Appendix~\ref{app:pseudocode}, Alg.~\ref{alg:choose_pmds}). METRIK determines optimality by calculating the PMD's efficiency and the imputer's performance. For continuous metrics, we report the normalized root-mean-square deviation (nRMSD) over masked elements \citep{luo2022evaluating} [see Eq.~(\ref{eq:nrmsd}) in Appendix~\ref{app:eval_metrics}], which we negate for notational convenience under the performance maximization setting. For categorical metrics, we calculate accuracy and macro F1 \citep{kelleher2020fundamentals} over masked elements per metric [for example, see Eq.~(\ref{eq:acc}) in Appendix~\ref{app:eval_metrics}] and, analogous to the averaging operation used for nRMSD, report the median of the scores across metrics to yield a single score per performance metric, which we refer to as pooled accuracy (pACC) and pooled macro F1 (pMF1). 

Given metrics for assessing optimality, METRIK compares the performance of each learned imputer-PMD pair ($m^*$, $pmd^*$) and each initial imputer-PMD pair ($m_{ref}$, $pmd_{ref}$) on the validation set of the pilot study dataset, i.e., $D_{pilot,val}$, and ranks the learned imputer-PMD pairs based on the performance differences and design objective. We rank these pairs based on performance differences since the set of initial imputer-PMD pairs serves as a reference for attainable operating points, i.e., pairs of sampling efficiency and imputation performance. To estimate performance differences accurately, given the small validation set, METRIK calculates confidence intervals \citep{rosner2015fundamentals} per performance metric and uses them to lowerbound the performance gain of the learned pair over the reference pair. Specifically, it calculates confidence intervals for each pair's performance, given by $p_{m^*}$ and $p_{m_{ref}}$, sets the performance of the learned pair to the lower limit of the confidence interval, given by $p_{m^*,lower}$, and estimates the performance of the reference pair as the upper limit of the confidence interval, given by $p_{m_{ref},upper}$. It also records the PMD's efficiency associated with each pair, given by $e_{m^*}$ and $e_{m_{ref}}$ for the learned and reference imputer-PMD pairs, respectively.    

Then, to identify optimal imputer-PMD pairs, METRIK ranks candidate pairs based on their improvements in efficiency and imputation performance relative to each reference imputer-PMD pair. Specifically, METRIK first identifies eligible candidate pairs, i.e., those with higher efficiency ($e_{m^*} > e_{m_{ref}}$) and imputation performance ($p_{m^*,lower} > p_{m_{ref},upper}$), and stores them in set $M^*_{eligible}$. Then, METRIK uses stable sorting to first sort the eligible candidate pairs by efficiency and then sort by imputation performance when the design objective is to maximize efficiency and \textit{vice versa} when the objective is to maximize imputation performance. It concludes by storing the top candidate imputer-PMD pair ($m^*_{top}$, $pmd^*_{top}$) per reference imputer-PMD pair in solution set $M_{sol}$. If eligible candidates do not exist, the algorithm stores the reference pair instead. 

Given $M_{sol}$, the user chooses the imputer-PMD pair with desired performance characteristics and uses it during the execution of the Phase-3 trial. Specifically, the Phase-3 study only collects measurements based on the PMD and then imputes the corresponding dropped measurements using the imputer.


 \section{Experimental Setup} \label{sec:exp_setup}

In this section, we describe the datasets, baseline methods, and performance measures used for our experiments. Implementation details (e.g., hyperparameter selection) are  given in Appendix~\ref{app:impl_details}.

\subsection{Datasets}

We evaluate METRIK on five real-world clinical RCT datasets, which we obtained from NINDS \citep{ninds}. These RCTs are Phase-2/Phase-3 trials that compare the effect of 1-2 experimental drugs against a control condition or standard-of-care in treating various neurological disorders, e.g., myasthenia gravis. For each RCT, we extract a dataset (details of our convenience sample generation can be found in Appendix~\ref{app:datasets:csg}), whose characteristics, i.e., the total number of subjects $n_{s,total}$, number of visits during trial $n_t$, number of metrics $n_m$, along with the percentage of measurements labeled as part of the trial protocol and percentage of metrics that are continuous are reported for each dataset in Table~\ref{tab:datasets}. For all datasets, we set $n_{s,pilot}$ to 60, the recommended pilot study size for standard RCTs \citep{teare2014sample}. The remainder of the dataset is used for testing.   

\begin{table}
  \caption{Dataset Properties}
  \label{tab:datasets}
  \centering
  \begin{tabular}{lccc}
    \toprule
    RCT (PI) Name & ($n_{s,total}$, $n_t$, $n_m$) & \% part of protocol & \% continuous \\
    \midrule
    CEF (Merit E. Cudkowicz) \citep{cudkowicz2014safety} & (448,37,457) & 3\% & 55\% \\
    ICARE (Carolee J. Winstein) \citep{winstein2016effect} & (361,5,560) & 22\% & 17\% \\
    FSZONE (Karl Kieburtz) \citep{neurol2015pioglitazone} & (210,6,187) & 28\% & 12\% \\
    MGTX (Gary Cutter) \citep{wolfe2016randomized} & (125,18,348) & 28\% & 36\% \\ 
    NN102 (Robert J. Fox) \citep{fox2018phase} & (255,13,213) & 27\% & 48\% \\
    \bottomrule
  \end{tabular}
\end{table}


\subsection{Baseline Algorithms}

We compare PMDs learned by METRIK against those generated by several \textit{ad hoc} strategies \citep{rioux2020reflection}. We implement a matrix sampling design, in which each measurement is randomly sampled based on a Bernoulli distribution. We refer to this baseline as RSD. We also implement the MF design and its variant called the Multiform Longitudinal (MFL) design, which randomizes form assignments over time across study participants. We also make comparisons against the Wave design \citep{graham2001planned}, which randomly samples the timepoints (rather than the metrics) at which measurements are taken. Specifically, we consider one version that includes the endpoints from the study (Wave+) and another version that excludes the endpoints (Wave), given that prior work demonstrates the advantage of this design \citep{rioux2020reflection}. Additional implementation details are provided in Appendix~\ref{app:impl_details:baseline}. By design, certain PMD generation strategies are not able to achieve specific baseline efficiency levels; the MF and MFL designs have a minimum efficiency of about 30\% \citep{graham2006planned} while the Wave designs are restricted by the total number of timepoints for a given RCT. 

\subsection{Performance Evaluation}

We report changes in efficiency ($e$) and imputation performance (nRMSD, pACC, and pMF1) relative to the baseline algorithms.  We evaluate performance using baseline efficiencies of \{5\%, 10\%, 30\%, 50\%, 70\%, 90\%\}, perform 5-fold cross validation per dataset, and show the distribution of the performance measures on the test set across the folds and datasets using boxplots. To denoise performance estimates under the random sampling designs, we sample 10K masked elements from the test set. We also visualize the PMDs to analyze the masking patterns learned by METRIK. 

\section{Results and Discussion} \label{sec:results}

First, we present results that demonstrate the performance advantage of METRIK over the baseline algorithms. Then, we present results from ablation studies that demonstrate the key ingredients underlying METRIK's effectiveness.

\subsection{METRIK vs. Baselines}

METRIK outperforms the baselines based on random sampling across different efficiency levels and under both design objectives: maximize efficiency or maximize imputation performance. 

\begin{figure} 
    \centering
    \includegraphics[width=\textwidth]{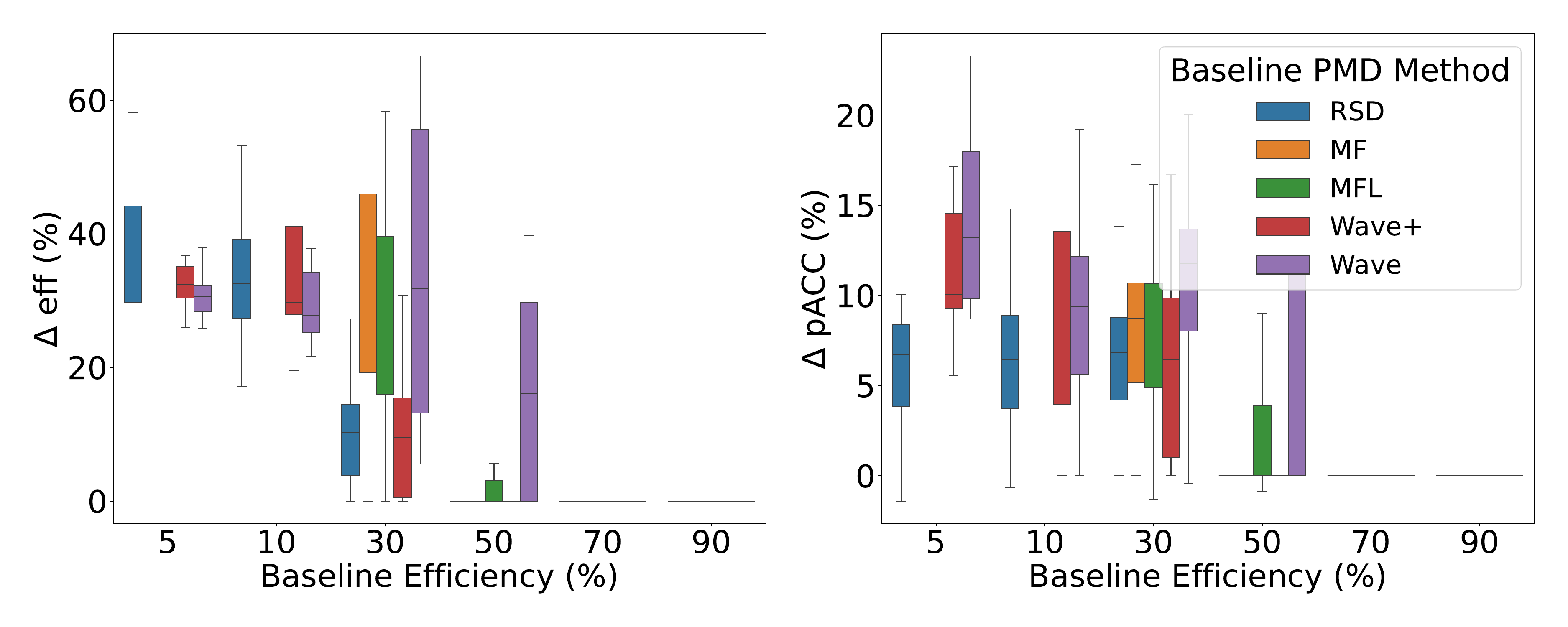}
    \caption{Performance gains obtained by METRIK over baseline PMD algorithms for a setting that maximizes efficiency. For MF and MFL, PMDs are only feasible for efficiency levels $\geq$ 30\%.}
    \label{fig:me_cat}
\end{figure}

Fig.~\ref{fig:me_cat} presents results for a design that aims to increase sampling efficiency over categorical metrics (changes in pMF1 are omitted for visual simplicity and are shown in Fig.~\ref{fig:me_cat_comp}). Compared to RSD with 5\% efficiency, METRIK increases efficiency by a median of 38\% (IQR: [30\%,44\%]) while improving performance (median gain in pACC is 7\%, IQR: [4\%, 8\%] and median gain in pMF1 is 4\%, IQR: [0,13\%]). METRIK outperforms RSD across higher baseline efficiencies, although efficiency gains diminish (e.g., efficiency increases by a median of 10\%, IQR: [4\%,14\%] relative to RSD with 30\% efficiency) because the maximum achievable efficiency is 100\%. In addition, since categorical metrics tend to dominate the dataset, we hypothesize that at higher baseline masking efficiencies, more uncorrelated metrics (e.g., metrics across different forms) are necessarily sampled, thereby reducing imputation performance and making it difficult for METRIK to find better PMDs. In these settings, METRIK performs no worse than the baseline because it chooses the baseline PMD. 

METRIK also outperforms other baseline algorithms, although gains in efficiency can be higher than those for RSD. For example, when the baseline efficiency is 30\%, METRIK increases efficiency by a median of 10\%, IQR: [4\%,14\%] for RSD, 29\% (IQR: [19\%,46\%]) for MF, 22\% (IQR: [16\%,40\%]) for MFL, 9\% (IQR: [1\%,15\%]) for Wave+, and 32\% (IQR: [13\%, 56\%]) for Wave (all methods generally improve pACC and pMF1). We hypothesize that gains across other baseline algorithms are comparable or higher because these algorithms sample a subset of random designs, making it difficult for the imputer to learn more complex interactions across measurements. 

Similar performance gains under METRIK manifest over continuous metrics, as shown in Fig.~\ref{fig:me_con}. 

\begin{figure} 
    \centering
    \includegraphics[width=\textwidth]{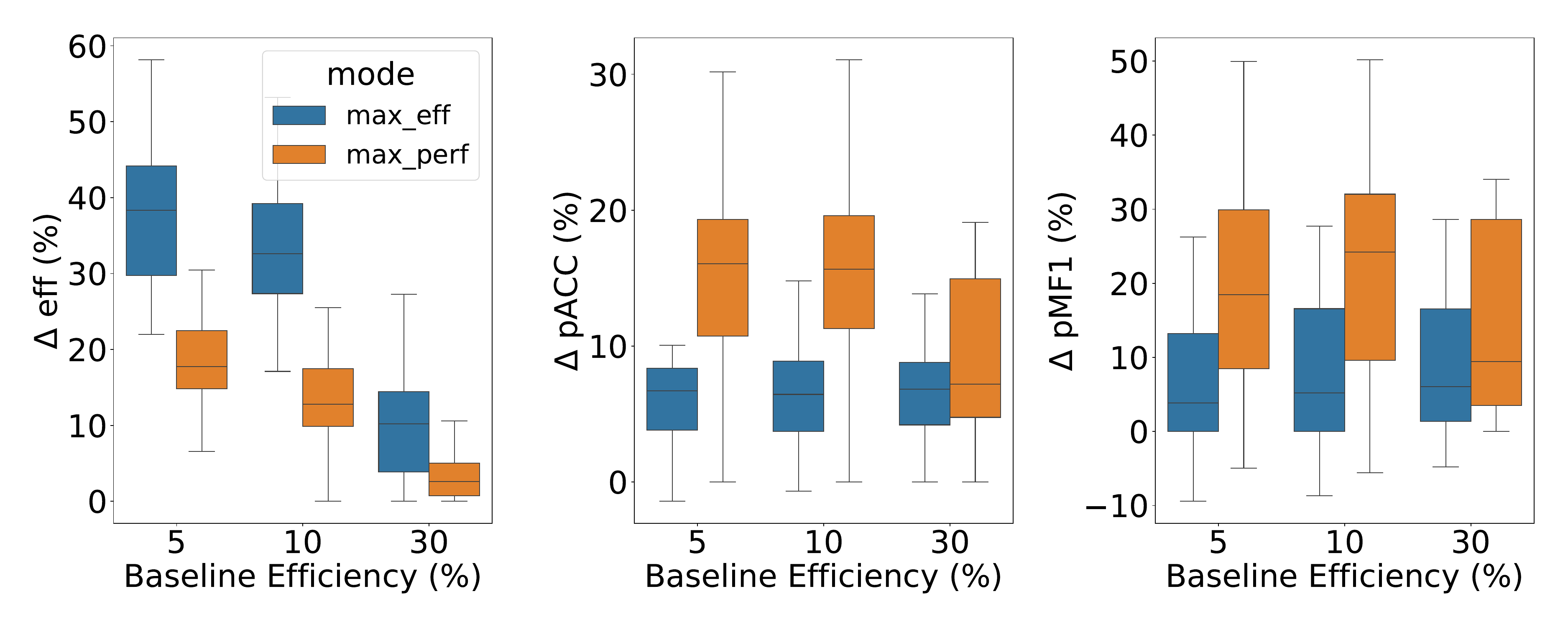}
    \caption{Performance gains obtained by METRIK over the RSD baseline under the two design objectives, i.e., maximize efficiency or maximize imputation performance. Results are only shown for baseline efficiencies ranging from 5-30\% since METRIK neither improves nor hurts performance at higher baseline efficiencies.}
    \label{fig:mode_compare}
\end{figure}

For a design that aims to maximize imputation performance at a given efficiency level, METRIK produces PMDs that appropriately trade off efficiency with imputation performance. Fig.~\ref{fig:mode_compare} compares the modes for the RSD baseline (we only show results for the RSD baseline since similar trends hold across other baselines). For example, at a baseline efficiency of 5\%, the version of METRIK that maximizes performance increases efficiency by a median of 18\% (IQR: [15\%,22\%]), pACC by a median of 16\% (IQR: [11\%,19\%]), and pMF1 by a median of 18\% (IQR: [8\%,30\%]), while the version of METRIK that maximizes efficiency increases efficiency by a median of 38\% (IQR: [30\%, 44\%]), pACC by a median of 7\% (IQR: [4\%,8\%]), and pMF1 by a median of 4\% (IQR: [0\%,13\%]). Similar tradeoffs hold across higher baseline efficiencies and over continuous metrics (Fig.~\ref{fig:mode_compare_con}). 

\begin{figure}
    \centering
    \includegraphics[width=\textwidth]{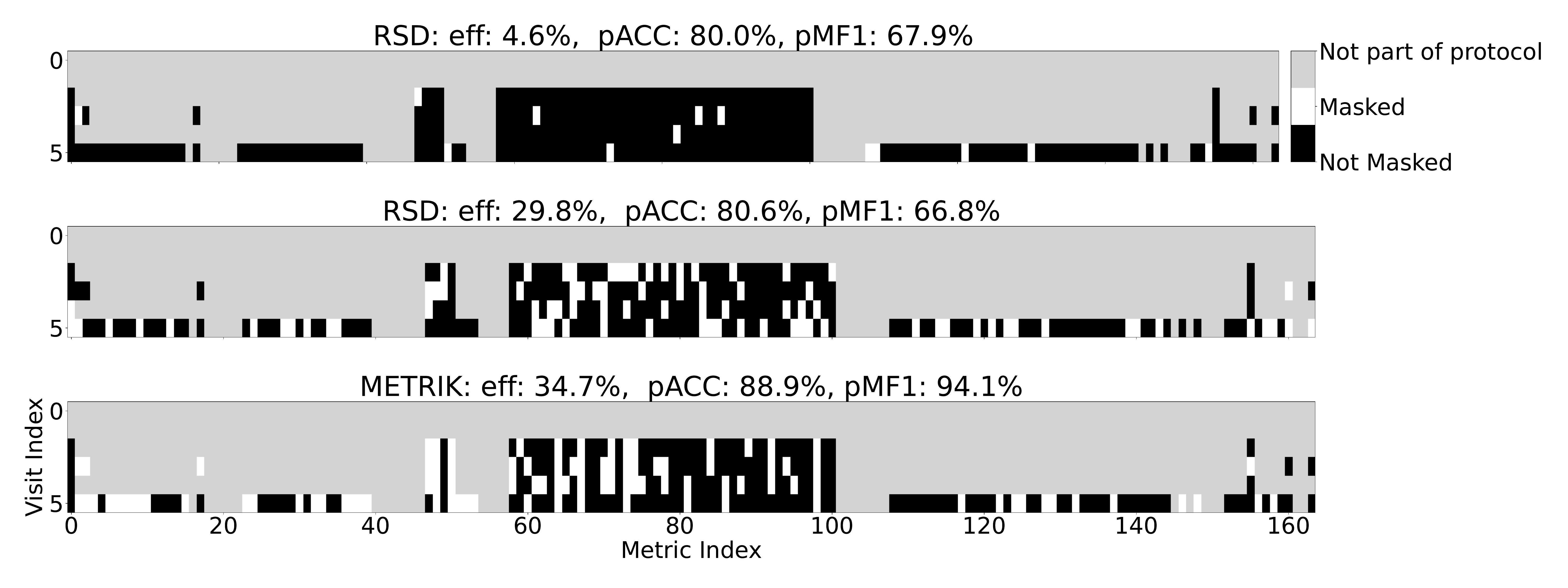}
    \caption{Sample PMDs produced by RSD (top two rows) and a sample PMD generated by METRIK (bottom row) on the FSZONE dataset for a setting that maximizes efficiency.}
    \label{fig:sample_mask}
\end{figure}

METRIK's ability to increase efficiency and imputation performance stems from its ability to identify correlated measurements over time and across metrics. To assess this, we compare the masking patterns under RSD-based PMDs against those under METRIK. Specifically, Fig.~\ref{fig:sample_mask} shows two RSD-generated PMDs along with one generated by METRIK on the FSZONE dataset when maximizing for efficiency, where the RSD-based PMDs have worse efficiency and imputation performance compared to the METRIK-based PMD. Compared against the RSD-based PMDs, the  METRIK-based PMD is less fragmented and masks out longer sequences along the metric or temporal dimensions. For example, it masks out metrics indexed 6 through 11 and 51 through 54. These ranges correspond to metrics from the same form. Specifically, 6-11 are assessments of abnormalities of the cranial nerves and 51-54 are assessments of abnormalities in muscle strength, both conducted in a neurological exam. METRIK also masks out time segments for certain metrics; for example, indices 70-75, which assess the patient's capacity to perform daily activities, are masked out for most visits past the baseline visit. Similar patterns emerge among METRIK-based PMDs for other RCT datasets and metric types (Fig.~\ref{fig:sample_mask_more}), demonstrating that METRIK leverages correlations over time and metrics to improve the quality of the PMD.

\subsection{Ablation Studies}

To understand key components underlying METRIK's effectiveness, we conduct ablations on the PMD selection step since design choices underlying the initial imputation fitting and PMD learning steps, , e.g., mask weight initialization, are based on tuning. For PMD selection, we investigate the effect of two design decisions: the algorithm used to generate candidate PMDs and the use of confidence intervals in estimating the imputation performance associated with a given PMD. 

\subsubsection{Source for Candidate PMDs}
\begin{figure}
    \centering
    \includegraphics[width=\textwidth]{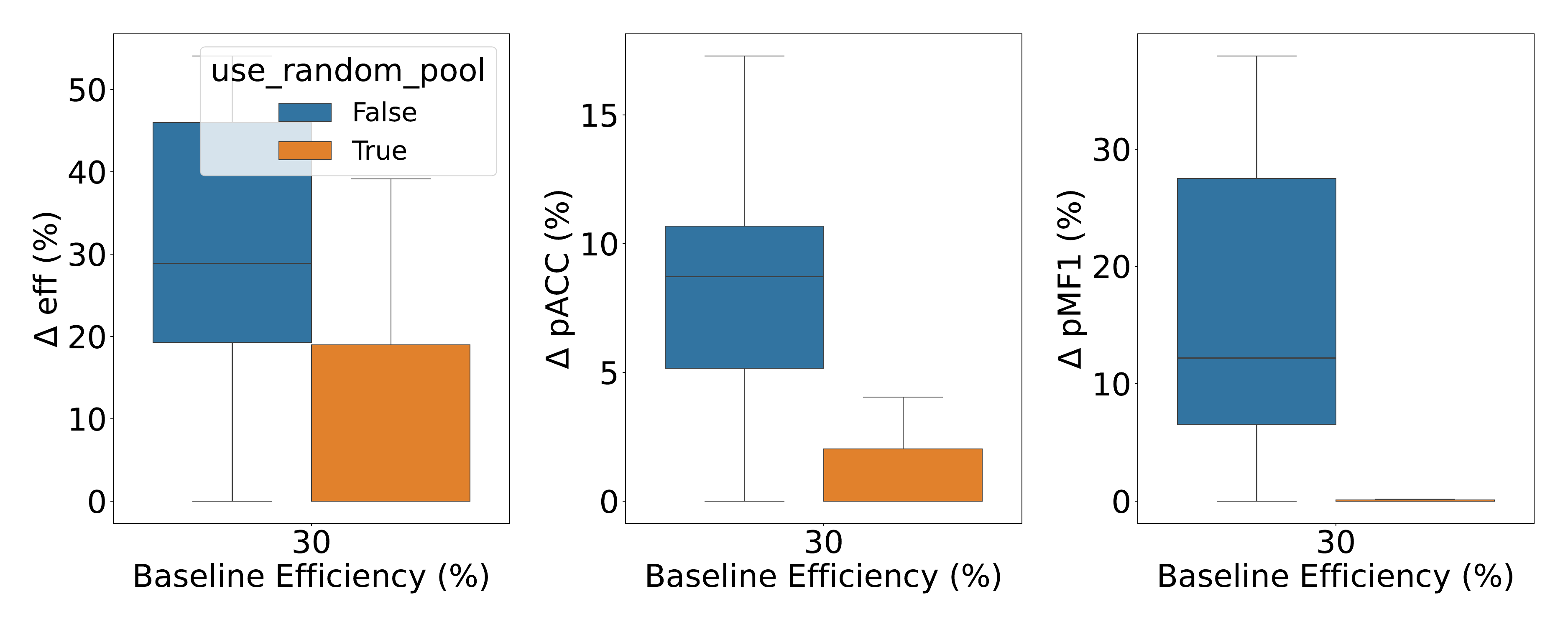}
    \caption{Performance gains under METRIK and an ablated version that replaces the candidate pool of learned PMDs with random ones generated by the MF design. Gains are measured with respect to an MF-based PMD baseline for a setting that maximizes efficiency. Results are shown for 30\% baseline efficiency, as the ablated method shows no gains at other efficiencies.}
    \label{fig:ablation_random_pool_cat}
\end{figure}

Instead of using PMDs generated through mask weight learning, we fill the candidate pool with PMDs generated by the random sampling baselines to demonstrate that random sampling is unlikely to yield solutions since the masked metrics are less correlated. Fig.~\ref{fig:ablation_random_pool_cat} shows results from this ablation for categorical metrics when the trial objective is to maximize efficiency. For simplicity, we only report results under the MF design at a baseline efficiency of 30\% since the MFL design gives comparable results and the remaining designs show no improvements across baseline efficiencies. Compared to METRIK, the ablated framework yields poorer solutions. At a baseline efficiency of 30\%, using a random candidate pool based on the MF design improves efficiency over an MF-based PMD by a median of 0\% (IQR: [0\%,19\%]) compared to METRIK, which improves efficiency over an MF-based PMD by 29\% (IQR: [19\%, 46\%]) while obtaining larger gains in pACC and pMF1. The performance drop under the ablated framework extends to continuous metrics (Fig.~\ref{fig:ablation_random_pool_con}) and the performance maximization setting of METRIK (Fig.~\ref{fig:ablation_random_pool_cat_mp}-~\ref{fig:ablation_random_pool_con_mp}).      
 
\subsubsection{Confidence Intervals for Performance Estimation}

\begin{figure} 
    \centering
    \includegraphics[width=\textwidth]{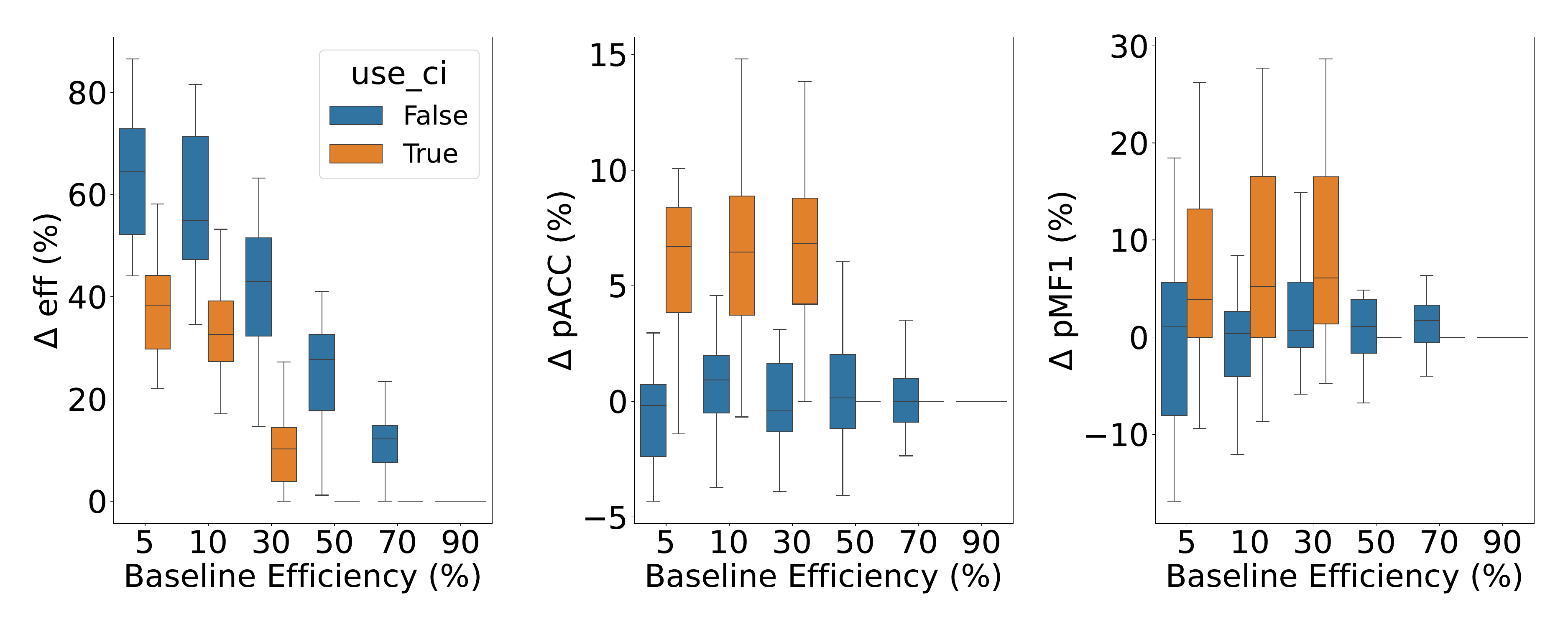}
    \caption{Performance gains under METRIK and an ablated version that does not use confidence intervals for performance estimation during PMD selection. Gains are measured with respect to the RSD baseline for a setting that maximizes efficiency.}
    \label{fig:ci_ablation_cat}
\end{figure}

We demonstrate that using confidence intervals improves METRIK's performance by removing them from the performance estimation step. Results from this ablation for categorical metrics under the efficiency maximization setting are shown in Fig.~\ref{fig:ci_ablation_cat}, where we show performance changes relative to the RSD baseline since the RSD baseline is competitive among the other random sampling baselines. At 5\% baseline efficiency, METRIK, with and without confidence intervals, boosts efficiency by a median of 38\% (IQR: [30\%,44\%]) and 64\% (IQR: [52\%,73\%]), respectively. However, METRIK without confidence intervals achieves higher efficiency by hurting imputation performance. Specifically, it degrades pACC by a median of 0\% (IQR: [-2\%,1\%]) and degrades pMF1 (median change in pMF1 is 1\%, IQR: [-8\%,6\%]), while METRIK does not (median change in pACC is 7\%, IQR: [4\%,8\%] and median change in pMF1 is 4\%, IQR: [0\%,13\%]). Similar effects from this ablation are observed at higher baseline efficiencies, as well as for continuous metrics (Fig.~\ref{fig:ci_ablation_con}) and under the performance maximization setting (Fig.~\ref{fig:ci_ablation_cat_mp}-~\ref{fig:ci_ablation_con_mp}). The ablated framework shows that using confidence intervals is an effective strategy at denoising performance estimates obtained on small validation datasets, which thereby enables METRIK to yield solutions with better performance guarantees.     

\section{Limitations and Future Work} \label{sec:limitations}
Having demonstrated METRIK's effectiveness, next, we discuss its limitations and future work. One shortcoming is that METRIK may compromise performance over certain metrics while improving aggregate performance across the metrics. This performance differential can be problematic if the compromised metrics are more important for the study. The safeguard against this shortcoming is for users of METRIK to analyze the results at a per-metric level (on the validation set), to ensure that changes in efficiency/errors over metrics of interest are in a favorable direction. Future work can also incorporate additional user constraints into METRIK to mitigate this issue. Another issue is that certain operating points may not be feasible to attain (e.g., high imputation performance) under imputation models trained on small datasets and will therefore require an increase in the training (pilot study) dataset size, but at the cost of reducing the total number of measurements that can be saved by the framework since fewer subjects will be subjected to the PMD. Future work will address this limitation by developing an adaptive strategy that increases the training dataset size to yield operating points with better imputation performance while accounting for the tradeoff in the total number of measurements collected across subjects. 

\section{Conclusion} \label{sec:conclusion}

To conclude, we have presented METRIK, a novel framework that learns a PMD for an RCT from a small pilot study by leveraging a state-of-the-art Transformer-based imputation framework with mask weight learning and then implementing a sampling strategy that generates diverse candidate PMDs and a selection strategy that chooses PMDs satisfying the design objective. Results across several real-world RCT datasets demonstrate that METRIK is effective at discovering PMD-imputer pairs with higher efficiency and better imputation performance compared to \textit{ad hoc} random sampling designs. Given its performance, METRIK can be used by trial designers to automatically design PMDs for RCTs without requiring additional data overhead, thereby solving the problem of improving the efficiency (costs) of the data collection process in RCTs without needing to remove any metric.        


\bibliographystyle{unsrtnat}
\bibliography{ref}

\begin{thebibliography}{34}
\providecommand{\natexlab}[1]{#1}
\providecommand{\url}[1]{\texttt{#1}}
\expandafter\ifx\csname urlstyle\endcsname\relax
  \providecommand{\doi}[1]{doi: #1}\else
  \providecommand{\doi}{doi: \begingroup \urlstyle{rm}\Url}\fi

\bibitem[Friedman et~al.(2015)Friedman, Furberg, DeMets, Reboussin, and Granger]{friedman2015fundamentals}
Lawrence~M. Friedman, Curt~D. Furberg, David~L. DeMets, David~M. Reboussin, and Christopher~B. Granger.
\newblock \emph{{Fundamentals of Clinical Trials}}.
\newblock Springer, 2015.

\bibitem[Sertkaya et~al.(2016)Sertkaya, Wong, Jessup, and Beleche]{sertkaya2016key}
Aylin Sertkaya, Hui-Hsing Wong, Amber Jessup, and Trinidad Beleche.
\newblock Key cost drivers of pharmaceutical clinical trials in the {United States}.
\newblock \emph{Clinical Trials}, 13\penalty0 (2):\penalty0 117--126, 2016.

\bibitem[Getz and Campo(2013)]{getz2013drug}
Kenneth Getz and Rafael Campo.
\newblock Drug development study designs have reached the danger zone.
\newblock \emph{Expert Review of Clinical Pharmacology}, 6\penalty0 (6):\penalty0 589--591, 2013.

\bibitem[Getz et~al.(2013)Getz, Kim, Stergiopoulos, and Kaitin]{getz2013new}
Kenneth~A. Getz, Jennifer Kim, Stella Stergiopoulos, and Kenneth~I. Kaitin.
\newblock New governance mechanisms to optimize protocol design.
\newblock \emph{Therapeutic Innovation \& Regulatory Science}, 47\penalty0 (6):\penalty0 651--655, 2013.

\bibitem[Rioux et~al.(2020)Rioux, Lewin, Odejimi, and Little]{rioux2020reflection}
Charlie Rioux, Antoine Lewin, Omolola~A. Odejimi, and Todd~D. Little.
\newblock Reflection on modern methods: Planned missing data designs for epidemiological research.
\newblock \emph{International Journal of Epidemiology}, 49\penalty0 (5):\penalty0 1702--1711, 2020.

\bibitem[Adig{\"u}zel and Wedel(2008)]{adiguzel2008split}
Feray Adig{\"u}zel and Michel Wedel.
\newblock Split questionnaire design for massive surveys.
\newblock \emph{Journal of Marketing Research}, 45\penalty0 (5):\penalty0 608--617, 2008.

\bibitem[Wu et~al.(2016)Wu, Jia, Rhemtulla, and Little]{wu2016search}
Wei Wu, Fan Jia, Mijke Rhemtulla, and Todd~D. Little.
\newblock Search for efficient complete and planned missing data designs for analysis of change.
\newblock \emph{Behavior Research Methods}, 48:\penalty0 1047--1061, 2016.

\bibitem[Imbriano(2018)]{imbriano2018methods}
Paul Imbriano.
\newblock \emph{{Methods for Improving Efficiency of Planned Missing Data Designs}}.
\newblock PhD thesis, University of Michigan, 2018.

\bibitem[Zerveas et~al.(2021)Zerveas, Jayaraman, Patel, Bhamidipaty, and Eickhoff]{zerveas2021transformer}
George Zerveas, Srideepika Jayaraman, Dhaval Patel, Anuradha Bhamidipaty, and Carsten Eickhoff.
\newblock A {T}ransformer-based framework for multivariate time series representation learning.
\newblock In \emph{Proceedings of the ACM SIGKDD Conference on Knowledge Discovery \& Data Mining}, 2021.

\bibitem[Vaswani et~al.(2017)Vaswani, Shazeer, Parmar, Uszkoreit, Jones, Gomez, Kaiser, and Polosukhin]{vaswani2017attention}
Ashish Vaswani, Noam Shazeer, Niki Parmar, Jakob Uszkoreit, Llion Jones, Aidan~N. Gomez, Lukasz Kaiser, and Illia Polosukhin.
\newblock Attention is all you need.
\newblock In \emph{Proceedings of the Annual Conference on Neural Information Processing Systems}, 2017.

\bibitem[Csord\'as et~al.(2021)Csord\'as, van Steenkiste, and Schmidhuber]{csordas2021neural}
R\'obert Csord\'as, Sjoerd van Steenkiste, and J\"urgen Schmidhuber.
\newblock Are neural nets modular? {I}nspecting functional modularity through differentiable weight masks.
\newblock In \emph{Proceedings of the International Conference on Learning Representations}, 2021.

\bibitem[Raghunathan and Grizzle(1995)]{raghunathan1995split}
Trivellore~E. Raghunathan and James~E. Grizzle.
\newblock A split questionnaire survey design.
\newblock \emph{Journal of the American Statistical Association}, 90\penalty0 (429):\penalty0 54--63, 1995.

\bibitem[Graham et~al.(2006)Graham, Taylor, Olchowski, and Cumsille]{graham2006planned}
John~W. Graham, Bonnie~J. Taylor, Allison~E. Olchowski, and Patricio~E. Cumsille.
\newblock Planned missing data designs in psychological research.
\newblock \emph{Psychological Methods}, 11\penalty0 (4):\penalty0 323--343, 2006.

\bibitem[Van~Buuren(2018)]{van2018flexible}
Stef Van~Buuren.
\newblock \emph{{Flexible Imputation of Missing Data}}.
\newblock CRC Press, 2018.

\bibitem[Van~Buuren and Groothuis-Oudshoorn(2011)]{van2011mice}
Stef Van~Buuren and Karin Groothuis-Oudshoorn.
\newblock {MICE}: Multivariate imputation by chained equations in {R}.
\newblock \emph{Journal of Statistical Software}, 45:\penalty0 1--67, 2011.

\bibitem[Luo et~al.(2018)Luo, Szolovits, Dighe, and Baron]{luo20183d}
Yuan Luo, Peter Szolovits, Anand~S. Dighe, and Jason~M. Baron.
\newblock {3D-MICE}: Integration of cross-sectional and longitudinal imputation for multi-analyte longitudinal clinical data.
\newblock \emph{Journal of the American Medical Informatics Association}, 25\penalty0 (6):\penalty0 645--653, 2018.

\bibitem[Luo(2022)]{luo2022evaluating}
Yuan Luo.
\newblock Evaluating the state of the art in missing data imputation for clinical data.
\newblock \emph{Briefings in Bioinformatics}, 23\penalty0 (1), 2022.

\bibitem[Liu et~al.(2023)Liu, Li, Yuan, Ong, Ning, Xie, Saffari, Shang, Volovici, Chakraborty, et~al.]{liu2023handling}
Mingxuan Liu, Siqi Li, Han Yuan, Marcus Eng~Hock Ong, Yilin Ning, Feng Xie, Seyed~Ehsan Saffari, Yuqing Shang, Victor Volovici, Bibhas Chakraborty, et~al.
\newblock Handling missing values in healthcare data: A systematic review of deep learning-based imputation techniques.
\newblock \emph{Artificial Intelligence in Medicine}, 142, 2023.

\bibitem[Y{\i}ld{\i}z et~al.(2022)Y{\i}ld{\i}z, Ko{\c{c}}, and Ko{\c{c}}]{yildiz2022multivariate}
A.~Yark{\i}n Y{\i}ld{\i}z, Emirhan Ko{\c{c}}, and Aykut Ko{\c{c}}.
\newblock Multivariate time series imputation with {T}ransformers.
\newblock \emph{IEEE Signal Processing Letters}, 29:\penalty0 2517--2521, 2022.

\bibitem[Teare et~al.(2014)Teare, Dimairo, Shephard, Hayman, Whitehead, and Walters]{teare2014sample}
M.~Dawn Teare, Munyaradzi Dimairo, Neil Shephard, Alex Hayman, Amy Whitehead, and Stephen~J. Walters.
\newblock Sample size requirements to estimate key design parameters from external pilot randomised controlled trials: A simulation study.
\newblock \emph{Trials}, 15:\penalty0 1--13, 2014.

\bibitem[Stanley(2007)]{stanley2007design}
Kenneth Stanley.
\newblock Design of randomized controlled trials.
\newblock \emph{Circulation}, 115\penalty0 (9):\penalty0 1164--1169, 2007.

\bibitem[Kelleher et~al.(2020)Kelleher, Mac~Namee, and D'arcy]{kelleher2020fundamentals}
John~D. Kelleher, Brian Mac~Namee, and Aoife D'arcy.
\newblock \emph{Fundamentals of Machine Learning for Predictive Data Analytics: Algorithms, Worked Examples, and Case Studies}.
\newblock The MIT Press, 2020.

\bibitem[Rosner(2016)]{rosner2015fundamentals}
Bernard Rosner.
\newblock \emph{Fundamentals of Biostatistics}.
\newblock Cengage Learning, 2016.

\bibitem[of~Neurological~Disorders and Stroke(2024)]{ninds}
National~Institute of~Neurological~Disorders and Stroke.
\newblock Archived clinical research datasets.
\newblock \url{https://www.ninds.nih.gov/current-research/research-funded-ninds/clinical-research/archived-clinical-research-datasets}, February 2024.

\bibitem[Cudkowicz et~al.(2014)Cudkowicz, Titus, Kearney, Yu, Sherman, Schoenfeld, Hayden, Shui, Brooks, Conwit, et~al.]{cudkowicz2014safety}
Merit~E. Cudkowicz, Sarah Titus, Marianne Kearney, Hong Yu, Alexander Sherman, David Schoenfeld, Douglas Hayden, Amy Shui, Benjamin Brooks, Robin Conwit, et~al.
\newblock Safety and efficacy of ceftriaxone for amyotrophic lateral sclerosis: A multi-stage, randomised, double-blind, placebo-controlled trial.
\newblock \emph{The Lancet Neurology}, 13\penalty0 (11):\penalty0 1083--1091, 2014.

\bibitem[Winstein et~al.(2016)Winstein, Wolf, Dromerick, Lane, Nelsen, Lewthwaite, Cen, Azen, et~al.]{winstein2016effect}
Carolee~J. Winstein, Steven~L. Wolf, Alexander~W. Dromerick, Christianne~J. Lane, Monica~A. Nelsen, Rebecca Lewthwaite, Steven~Yong Cen, Stanley~P. Azen, et~al.
\newblock Effect of a task-oriented rehabilitation program on upper extremity recovery following motor stroke: {T}he {ICARE} randomized clinical trial.
\newblock \emph{Journal of Americal Medical Association}, 315\penalty0 (6):\penalty0 571--581, 2016.

\bibitem[in~{P}arkinson Disease (NET-PD) FS-ZONE~investigators(2015)]{neurol2015pioglitazone}
NINDS Exploratory~Trials in~{P}arkinson Disease (NET-PD) FS-ZONE~investigators.
\newblock Pioglitazone in early {P}arkinson's disease: {A} phase 2, multicentre, double-blind, randomised trial.
\newblock \emph{Lancet Neurology}, 14\penalty0 (8):\penalty0 795--803, 2015.

\bibitem[Wolfe et~al.(2016)Wolfe, Kaminski, Aban, Minisman, Kuo, Marx, Str{\"o}bel, Mazia, Oger, Cea, et~al.]{wolfe2016randomized}
Gil~I. Wolfe, Henry~J. Kaminski, Inmaculada~B. Aban, Greg Minisman, Hui-Chien Kuo, Alexander Marx, Philipp Str{\"o}bel, Claudio Mazia, Joel Oger, J.~Gabriel Cea, et~al.
\newblock Randomized trial of thymectomy in myasthenia gravis.
\newblock \emph{New England Journal of Medicine}, 375\penalty0 (6):\penalty0 511--522, 2016.

\bibitem[Fox et~al.(2018)Fox, Coffey, Conwit, Cudkowicz, Gleason, Goodman, Klawiter, Matsuda, McGovern, Naismith, et~al.]{fox2018phase}
Robert~J. Fox, Christopher~S. Coffey, Robin Conwit, Merit~E. Cudkowicz, Trevis Gleason, Andrew Goodman, Eric~C. Klawiter, Kazuko Matsuda, Michelle McGovern, Robert~T. Naismith, et~al.
\newblock Phase 2 trial of ibudilast in progressive multiple sclerosis.
\newblock \emph{New England Journal of Medicine}, 379\penalty0 (9):\penalty0 846--855, 2018.

\bibitem[Graham et~al.(2001)Graham, Taylor, and Cumsille]{graham2001planned}
John~W. Graham, Bonnie~J. Taylor, and Patricio~E. Cumsille.
\newblock \emph{New Methods for the Analysis of Change}, chapter~11.
\newblock American Psychological Association, 2001.

\bibitem[Liu et~al.(2019)Liu, Jiang, He, Chen, Liu, Gao, and Han]{liu2019variance}
Liyuan Liu, Haoming Jiang, Pengcheng He, Weizhu Chen, Xiaodong Liu, Jianfeng Gao, and Jiawei Han.
\newblock On the variance of the adaptive learning rate and beyond.
\newblock 2019.
\newblock URL \url{https://arxiv.org/pdf/1908.03265}.

\bibitem[Paszke et~al.(2019)Paszke, Gross, Massa, Lerer, Bradbury, Chanan, Killeen, Lin, Gimelshein, Antiga, et~al.]{paszke2019pytorch}
Adam Paszke, Sam Gross, Francisco Massa, Adam Lerer, James Bradbury, Gregory Chanan, Trevor Killeen, Zeming Lin, Natalia Gimelshein, Luca Antiga, et~al.
\newblock {PyTorch}: An imperative style, high-performance deep learning library.
\newblock In \emph{Proceedings of the Annual Conference on Neural Information Processing Systems}, 2019.

\bibitem[Zerveas(2021)]{mvts_impl}
George Zerveas.
\newblock {Multivariate Time Series Transformer Framework}.
\newblock \url{https://github.com/gzerveas/mvts_transformer}, 2021.

\bibitem[Csord\'as(2021)]{mask_impl}
R\'obert Csord\'as.
\newblock {Codebase for Inspecting Modularity of Neural Networks}.
\newblock \url{https://github.com/RobertCsordas/modules}, 2021.

\end{thebibliography}







\newpage 

\appendix

\begin{appendices}
\counterwithin{figure}{section}
\counterwithin{table}{section}

\section{Pseudocode} \label{app:pseudocode}

\begin{algorithm}[H] 
\caption{generate\_candidate\_pmds}
\label{alg:gen_cand}
\SetKwInOut{Input}{Input}
\SetKwInOut{Output}{Output}
\Input{$D_{pilot}$, $P$, $S$, $\lambda_{range}$, $\eta_{range}$}
\Output{$M^{*}$}
\DontPrintSemicolon
\BlankLine
$M$ = set()\;

\For{e in range(0,100\%)}{
    $m_e$ = MVTS($D_{pilot}$, $P$, $S$, $pmd_{RSD}(e)$)
    
    $pmd_{RSD,e}$ = $pmd_{RSD}(e)$.sample()
    
    $M$.add(($m_e$, $pmd_{RSD,e}$))\;
    
}

$M^*$ = set()\;

\For{$m_e$ in $M$}{
    mask\_weight = $pmd_{RSD}(e)$.get\_prob()\;
    
    \For{$\lambda_{wm}$, $\eta$ in combinations($\lambda_{range}$, $\eta_{range}$)}{
        masked\_model = ModuleList([MaskLayer(mask\_weight), $m_e$])\;
        
        masked\_model.train($\lambda_{wm}$, $\eta$, $D_{pilot}$, $P$, $S$)\;

        $m^*$ = masked\_model.imputer
        
        $pmd^*$ = masked\_model.mask\_layer
        
        $M^{*}$.add(($m^*$, $pmd^*$))\;
    }
    
}

\Return $M^{*}$\;
\end{algorithm}

\begin{algorithm}[H]
\SetKwInOut{Input}{Input}
\SetKwInOut{Output}{Output}

\Input{$M$, $M^*$, $D_{pilot,val}$,  $design\_obj$}
\Output{$M_{sol}$}

\DontPrintSemicolon
\caption{choose\_pmds}
\label{alg:choose_pmds}

$M_{sol}$ = set()\;

\For{($m_{ref},pmd_{ref}$) in $M$}{
        $p_{m_{ref}}$ = get\_ci\_perf($m_{ref}$, $pmd_{ref}$, $D_{pilot,val}$)\;

        $e_{m_{ref}}$ = get\_eff($pmd_{ref}$)

        perf\_container = PerfContainer() \Comment{contains lists storing model, PMD, performance, efficiency}

        \For{($m^*$, $pmd^*$) in $M^*$}{
            $p_{m^*}$ = get\_ci\_perf($m^*$, $pmd^*$, $D_{pilot,val}$)\;
            
            $e_{m^*}$ = get\_eff($pmd^*$)\;
            
            perf\_container.add(($m^*$,$pmd^*$,$p_{m^*,lower}$, $e_{m^*}$))\;
        }

        eligible\_candidates = (perf\_container.perf > $p_{m_{ref},upper}$) \& (perf\_container.eff > $e_{m_{ref}}$)\;

        \If{len(eligible\_candidates) == 0}{
            $M\_{sol}$.add(($m_{ref}$, $pmd_{ref}$))\;
            
            continue\;
        }

        \If{design\_obj == ``max\_eff"}{
            keys = [``perf", ``eff"] \Comment{last key is the primary sort key}\;
        }
        \Else{
            keys = [``eff", ``perf"]\;
        }

        $m^*_{top}$, $pmd^*_{top}$ = stable\_sort(perf\_container[eligible\_candidates], keys)[0]\;

        $M_{sol}$.add(($m^*_{top}$, $pmd^*_{top}$))\;
    }
\Return{$M_{sol}$}\;

\end{algorithm}

\section{Equations} \label{app:eval_metrics}

Eq.~(\ref{eq:nrmsd}) shows the formula for calculating nRMSD, which is based on prior work \cite{luo2022evaluating}, where $n_s$ is the number of subjects, $n_t$ is the number of timepoints, $n_m$ is the number of metrics, $\mathbbm{1}_{i,j,m}$ is an indicator stating whether the measurement associated with subject $i$ at timepoint $j$ for metric $m$ has been masked out, $\hat{y}_{i,j,m}$ is the model's prediction, $y_{i,j,m}$ is the ground truth, and $y_m$ is the matrix of measurements across subjects and timepoints. The formula presented here differs from that in \cite{luo2022evaluating} in that it normalizes across subjects to avoid cases of dividing by zero and averages across all metrics to report a single score.

\begin{equation} \label{eq:nrmsd}
    nRMSD=\sqrt{\frac{\sum_{(i,j,m) \in n_s \times n_t \times n_m} \mathbbm{1}_{i,j,m}(\frac{|\hat{y}_{i,j,m}-y_{i,j,m}|}{max(y_m)-min(y_m)})}
{\sum_{(i,j,m) \in n_s \times n_t \times n_m} \mathbbm{1}_{i,j,m}}}
\end{equation}

Eq.~(\ref{eq:acc}) shows the formula for calculating the accuracy per metric, adopting the same notation as used in Eq.~(\ref{eq:nrmsd}). For simplicity, we only show the calculation for accuracy but the same approach, i.e., measuring performance over the masked elements only, is used for calculating the macro F1 per metric.

\begin{equation} \label{eq:acc}
acc(m) = \frac{\sum_{(i,j) \in n_s \times n_t} \mathbbm{1}_{i,j,m}(\hat{y}_{i,j,m} == y_{i,j,m})}
{\sum_{(i,j) \in n_s \times n_t} \mathbbm{1}_{i,j,m}}
\end{equation}

\section{Implementation Details} \label{app:impl_details}

In this section, we review implementation details for the METRIK framework and the baseline algorithms, along with the computational resources used in our experiments. 

\subsection{METRIK} \label{app:impl_details:metrik}

Here, we describe hyperparameters used by the METRIK framework. 

We train the initial imputer using the MVTS algorithm, where the architecture consists of a Transformer encoder and linear decoder; specifically, we adopt a Transformer encoder with three encoder blocks and eight heads, with the dimensionality of the model set to 64 and dimensionality of the feed-forward layer set to 256, following similar configurations from the original work \citep{zerveas2021transformer}. For simplicity, we train imputers for continuous and categorical covariates separately, where imputers for continuous variables are trained using the mean squared error loss over masked elements while imputers for categorical variables are trained using the cross-entropy loss over masked elements. Each model is trained for 6K epochs using the RAdam optimizer \citep{liu2019variance} with the learning rate set to $1 \times 10^{-3}$ (continuous metrics) or $1 \times 10^{-4}$ (categorical metrics) with early checkpointing. 

During mask weight training, we set $\lambda_{range} = \{1 \times 10^{-7},1 \times 10^{-6},1 \times 10^{-5}\}$ and $\eta_{range} = \{0.1,0.5,1,5,10\}$ (we found these ranges sufficient to yield solutions based on the training/validation set), and otherwise adopt the same optimization settings as used for training the initial imputer, although we use the last checkpoint since the mask stabilizes towards the end of training.   

For PMD selection, we calculate two-sided 95\% confidence intervals using the bootstrap algorithm \citep{rosner2015fundamentals} and set the number of bootstraps to 1K since that was sufficient for intervals to converge.

\subsection{Baselines} \label{app:impl_details:baseline}

We describe the implementation for each baseline next.

We implement the RSD baseline by having it sample each measurement under protocol $P$ based on a Bernoulli distribution determined by the baseline efficiency level.

For the MF design, we vary the total number of item sets used to construct the forms (maximum set to 20). This parameter determines the efficiency of the forms. We divide the metrics randomly and equally across item sets, following a suggested strategy \citep{graham2006planned}, and then generate a set of forms by obtaining every pairwise combination over item sets. For a given form, measurements not included in the chosen pair of item sets are masked out. The MFL design differs from MF in that at each timepoint it randomly chooses a form among the ones generated under the MF design. 

For Wave designs, we vary the number of timepoints that can be dropped (the maximum is determined by $n_t$, the number of timepoints for a given RCT). This parameter determines the efficiency of the form. Given the number of timepoints to drop, we generate all resulting combinations over eligible timepoints and mask out measurements across all selected timepoints. For the Wave+ design, the set of eligible timepoints includes the endpoints while for the Wave design, the set of eligible timepoints excludes the endpoints.     

\subsection{Computing Environment} \label{app:impl_details:compute}

We conducted all experiments using one CPU core for model training and mask weight learning (per trained model) and five CPU cores for PMD selection, using 2.6 GHz Intel Skylake/2.8 GHz Intel Cascade Lake processors with a total memory of at most 6GB. Each experimental run (i.e., defined by a complete run of the algorithm for a given dataset and seed) requires at most 444 CPU hours (conservative estimate). The total compute time for our experiments across datasets and seeds, including the baseline evaluations, is at most 15K CPU hours. Our experiments require more compute than the experiments reported in the article due to additional runs stemming from hardware configuration tuning, hyperparameter tuning, etc.

\subsection{Software} \label{app:impl_details:sw}

All experiments are implemented using PyTorch \citep{paszke2019pytorch} and standard numerical Python packages. 

Code for the MVTS algorithm is obtained from \citep{mvts_impl} and is distributed under the MIT license; the copyright is under George Zerveas since Year 2021.  

Code for the mask learning algorithm is obtained from \citep{mask_impl} and is distributed under BSD 3-Clause "New" or "Revised" License; the copyright is under Robert Csordas since Year 2020.

\section{Datasets} \label{app:datasets}

Details of the data source and our process for obtaining a convenience sample from the dataset are described next.

 \subsection{Convenience Sample Generation} \label{app:datasets:csg}
Per dataset, we obtain a convenience sample from the raw clinical dataset by applying a sequence of steps. First, we remove any data collection forms that are not prescribed according to the trial protocol, comprised measurements that have no associated study visit, or are inconveniently formatted. Afterwards, we remove entries including duplicate rows, NaN indices, duplicate columns or metrics, subjects not assigned to any treatment group in the RCT, and timepoints or visit ids not part of the trial protocol. Next, we post-process the data by standardizing NaN placeholders, removing metrics with mixed data types (e.g., string and numerical), and removing metrics that are constant or entirely NaN. We automatically determine the type of each metric (i.e., continuous vs. categorical) based on the data type and range of values and also drop metrics that have very low variance to avoid issues with normalization. We also automatically determine $P$, the protocol mask, by identifying measurements (i.e., specific metric-timepoint pairs) with sufficiently low missingness rates. Subject masks $S$ are automatically determined based on NaN entries.

\subsection{Source} \label{app:datasets:source}

All datasets can be obtained from NINDS \citep{ninds} by filling out the data request form at the following link:  \url{https://www.ninds.nih.gov/sites/default/files/migrate-documents/sig_form_revised_508c.pdf}. No license/copyright information is provided. The terms of use for the datasets require that the dataset be used for research purposes, that results from the research be published, that the dataset not be shared without permission from NINDS, and that NINDS, along with the study publication and trial investigators associated with the dataset, be acknowledged in the publication.

\section{Figures}

\begin{figure}[H]
\begin{subfigure}[H]\textwidth
\centering
    \includegraphics[width=\textwidth]{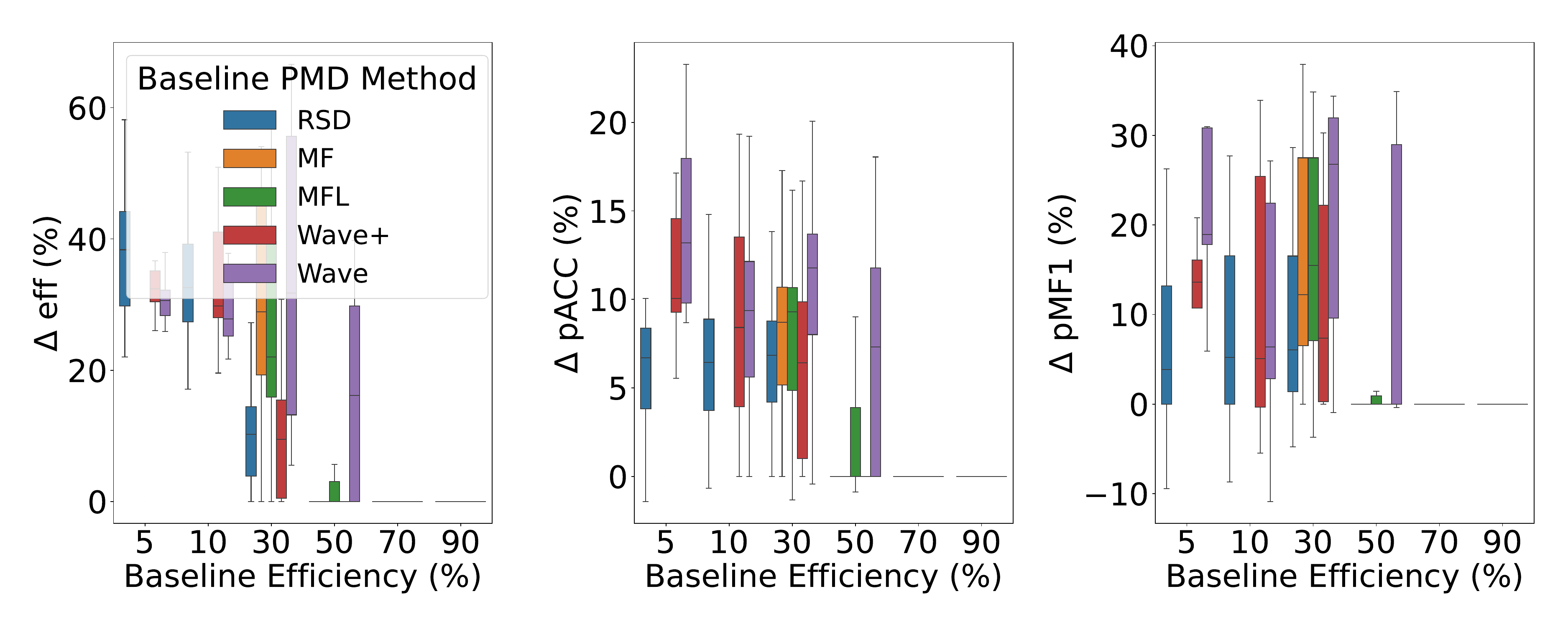}
    \caption{Complete results (i.e., pMF1 included) over categorical metrics. }
    \label{fig:me_cat_comp}
\end{subfigure}

\begin{subfigure}[H]\textwidth
    \centering
    \includegraphics[width=\textwidth]{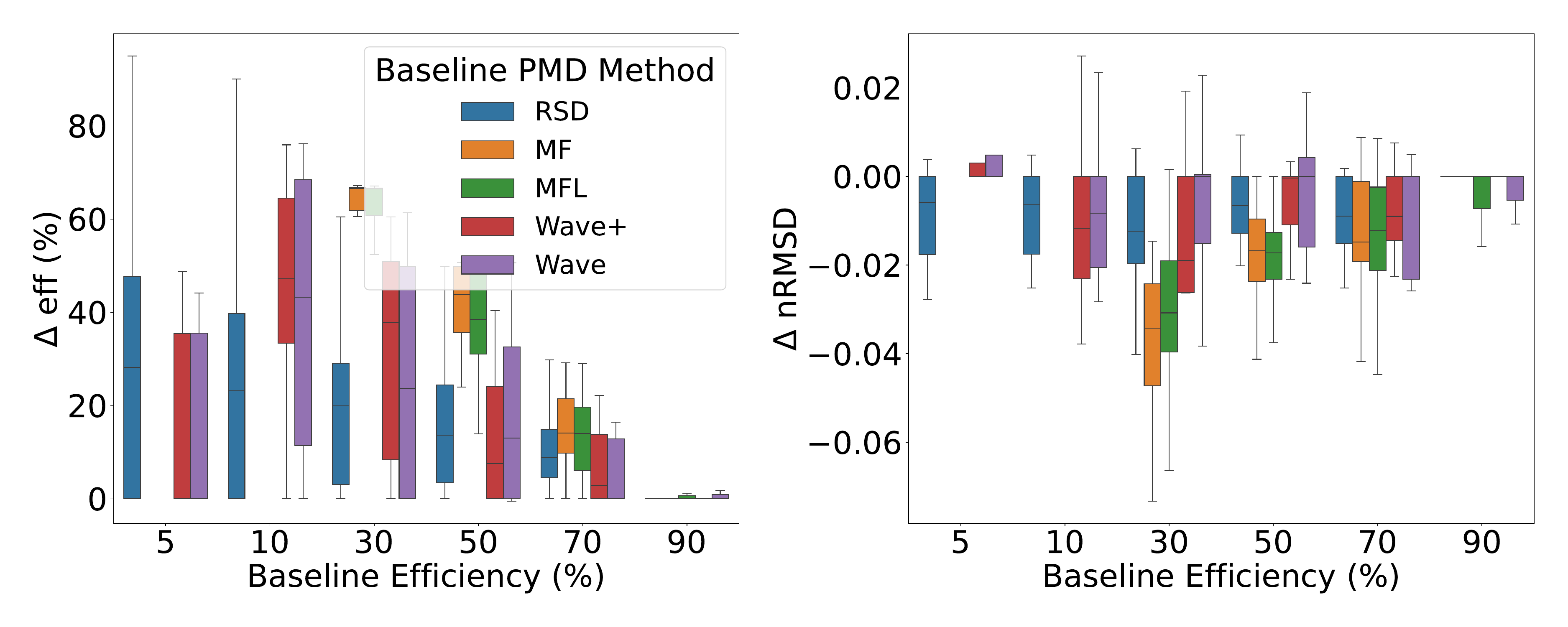}
    \caption{Results over continuous metrics.}
    \label{fig:me_con}
\end{subfigure}

\caption{Performance gains under METRIK over the baseline PMD algorithms across baseline efficiency levels for a setting where the design objective is set to maximize efficiency. MF and MFL designs are only possible for efficiencies $\geq$ 30\%.}
\label{fig:me}

\end{figure}

\begin{figure}[H]
    \centering
    \includegraphics[width=\textwidth]{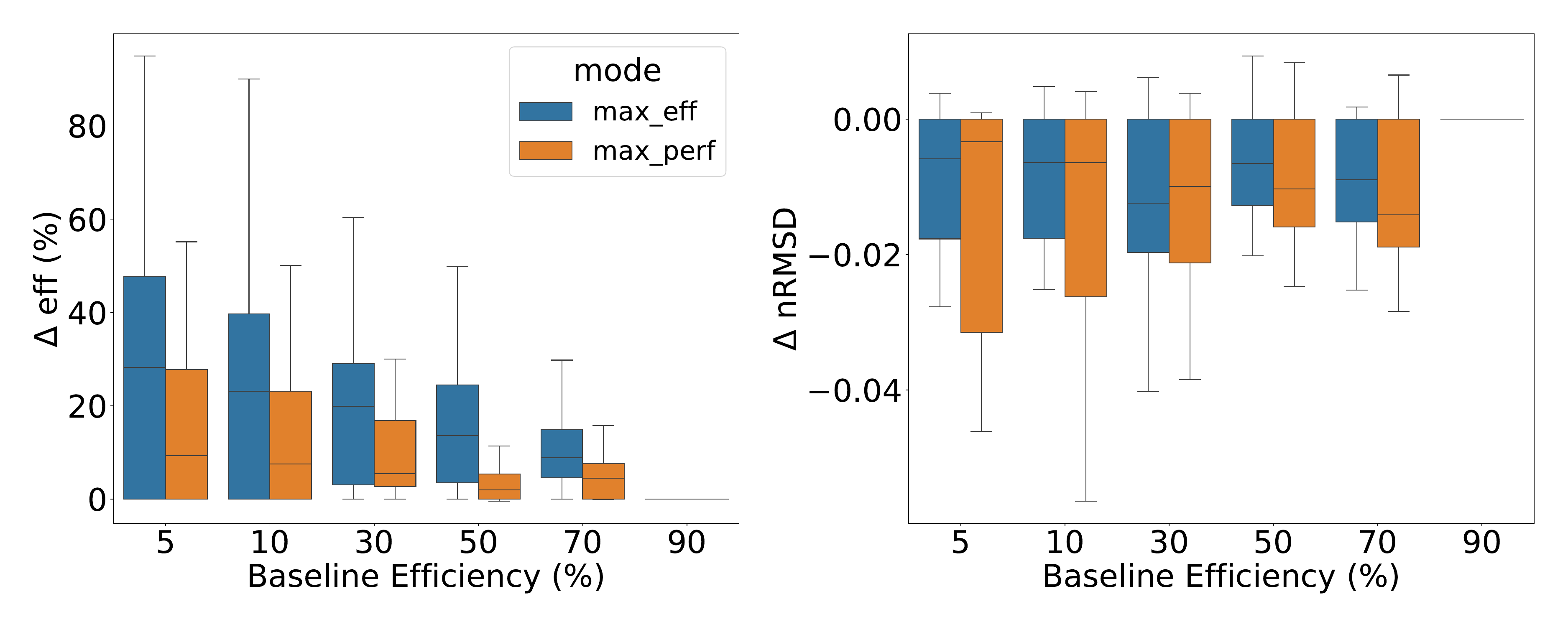}
    \caption{Results comparing the effect of the two design objectives, i.e., maximize efficiency or maximize imputation performance, on performance gains obtained by METRIK over the RSD baseline for a setting involving continuous metrics.}
    \label{fig:mode_compare_con}
\end{figure}

\begin{figure}[H]
    \centering
    \begin{subfigure}[b]{\textwidth}
        \centering
        \includegraphics[width=0.95\textwidth]{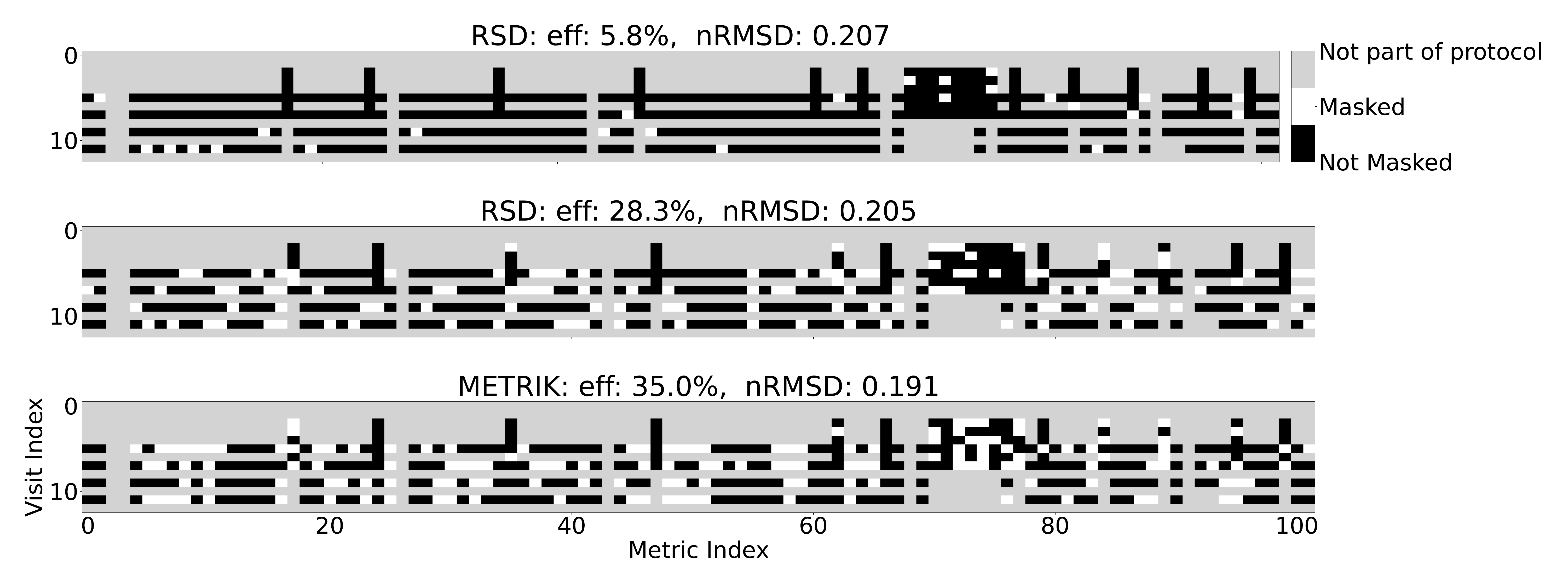}
        \caption{RSD PMDs (top two rows) and a METRIK PMD (bottom row) for continuous metrics on the NN102 dataset.}
         \label{fig:sample_mask_nn102}
    \end{subfigure}
    
    \begin{subfigure}[b]{\textwidth}
        \centering
        \includegraphics[width=0.95\textwidth]{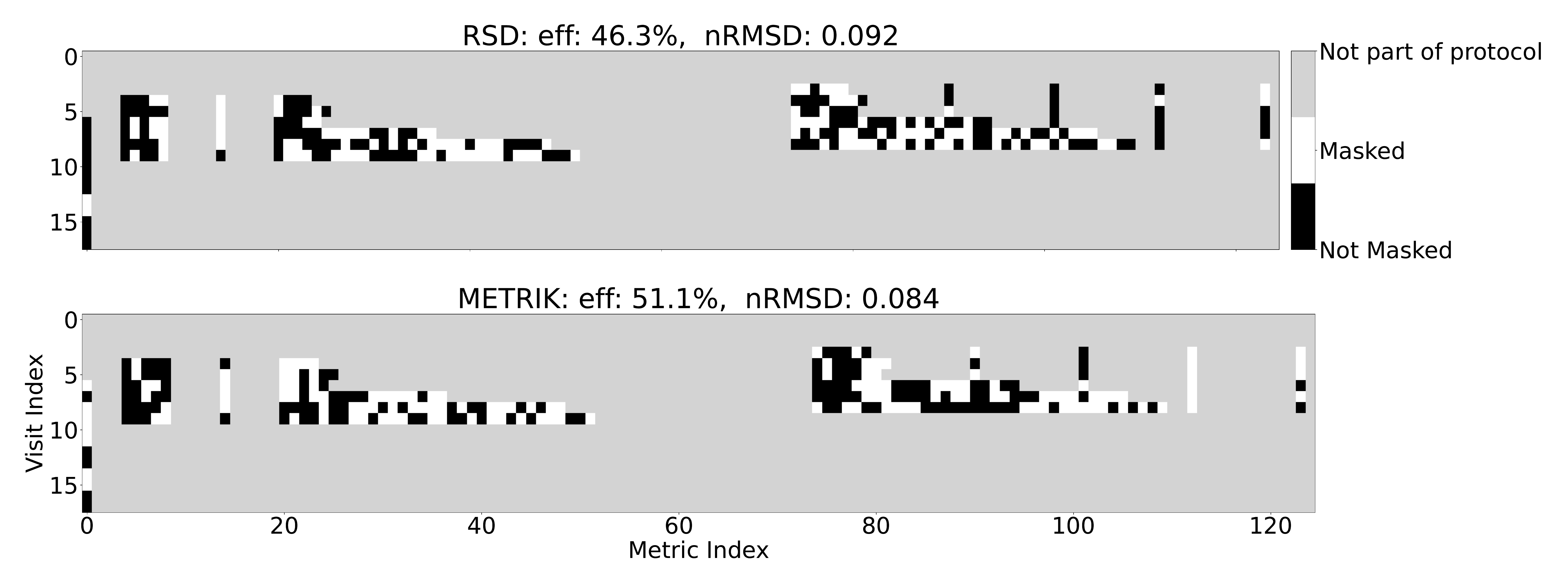}
        \caption{RSD PMDs (top row) and a METRIK PMD (bottom row) for continuous metrics on the MGTX dataset.}
         \label{fig:sample_mask_mgtx}
    \end{subfigure}

    \begin{subfigure}[b]{\textwidth}
        \centering
        \includegraphics[width=0.95\textwidth]{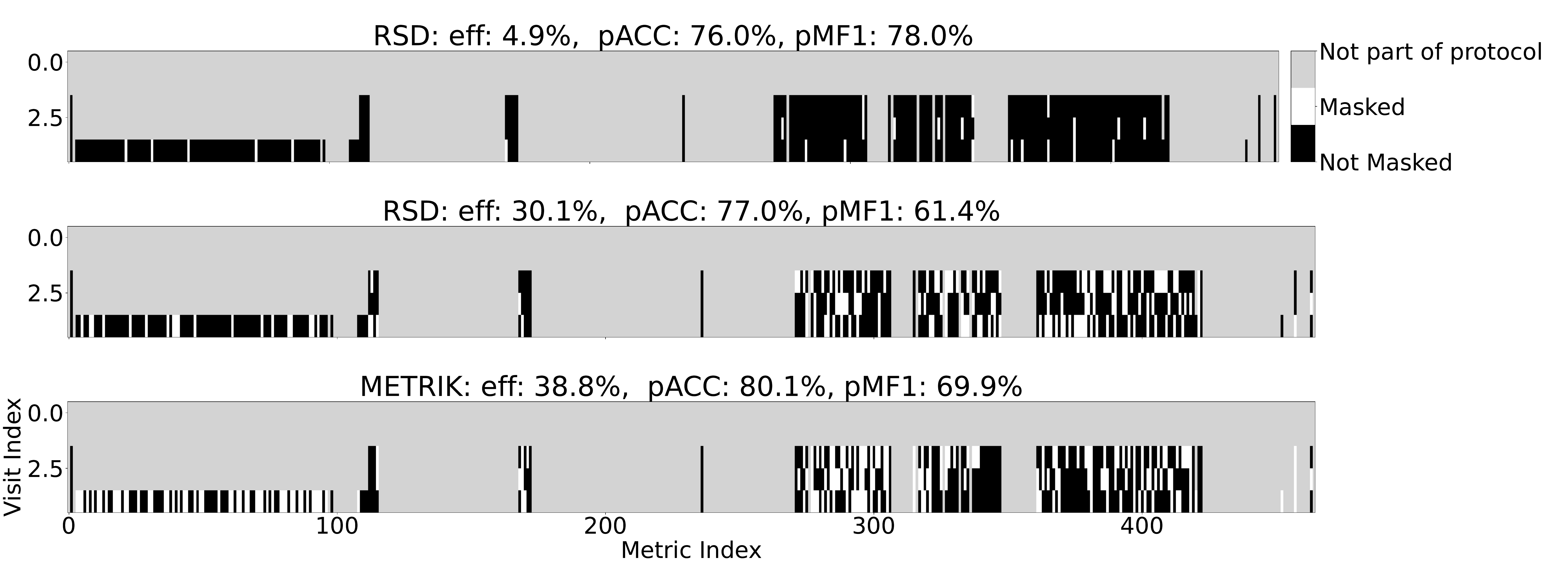}
        \caption{RSD PMDs (top two rows) and a METRIK PMD (bottom row) for categorical metrics on the ICARE dataset.}
         \label{fig:sample_mask_icare}
    \end{subfigure}

    \begin{subfigure}[b]{\textwidth}
        \centering
        \includegraphics[width=0.95\textwidth]{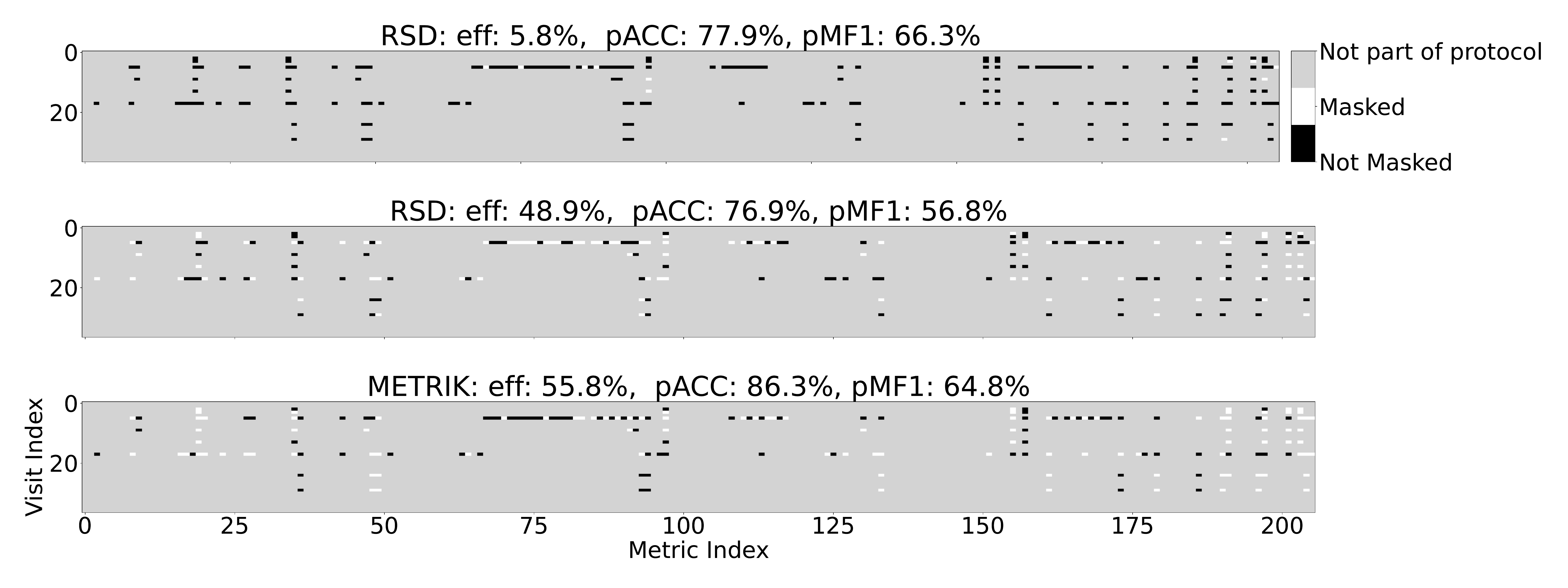}
        \caption{RSD PMDs (top two rows) and a METRIK PMD (bottom row) for categorical metrics on the CEF dataset.}
         \label{fig:sample_mask_cef}
    \end{subfigure}
    
    \caption{Sample PMDs produced by RSD and METRIK across datasets and different metric types for a setting where the design objective is set to maximize efficiency.}
    \label{fig:sample_mask_more}
\end{figure}

\begin{figure}[H]
    \centering
    \begin{subfigure}[b]{\textwidth}
        \centering
        \includegraphics[width=\textwidth]{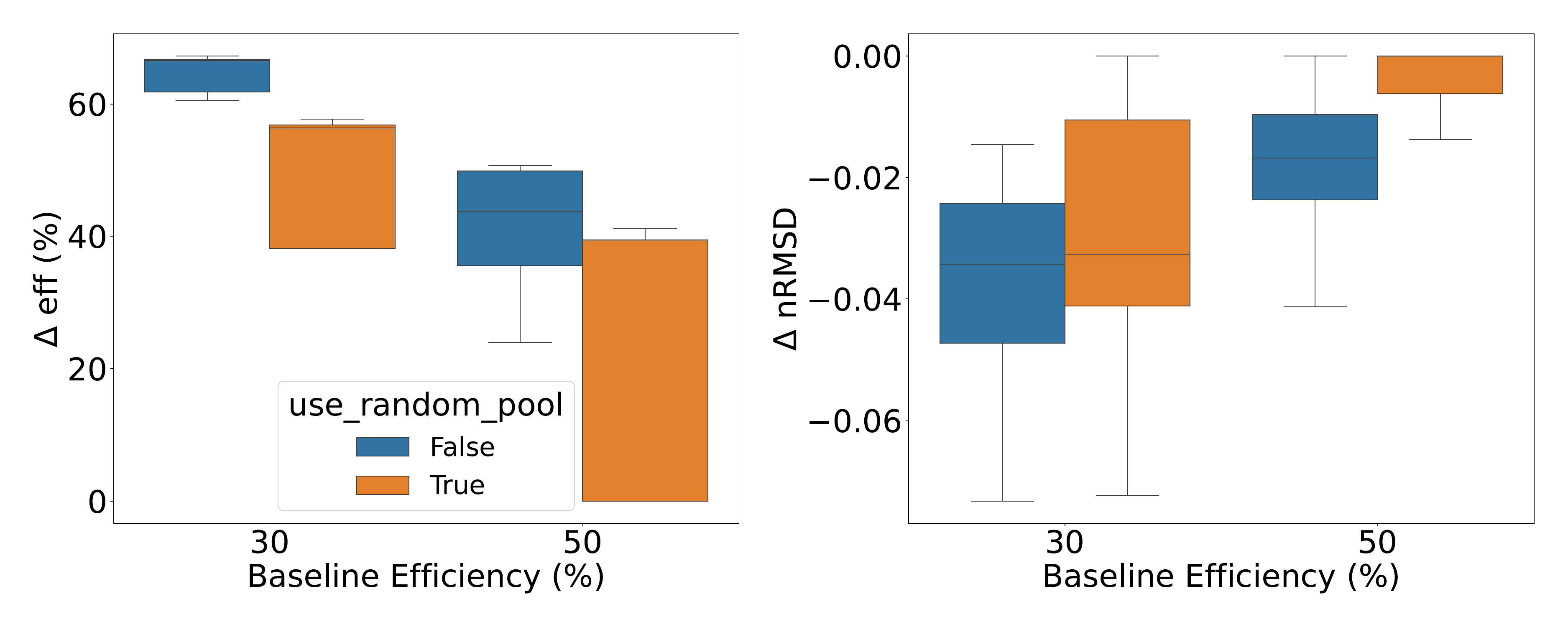}
        \caption{Results over continuous metrics for a setting where the design objective is set to maximize efficiency.}
         \label{fig:ablation_random_pool_con}
    \end{subfigure}

    \begin{subfigure}[b]{\textwidth}
        \centering
        \includegraphics[width=\textwidth]{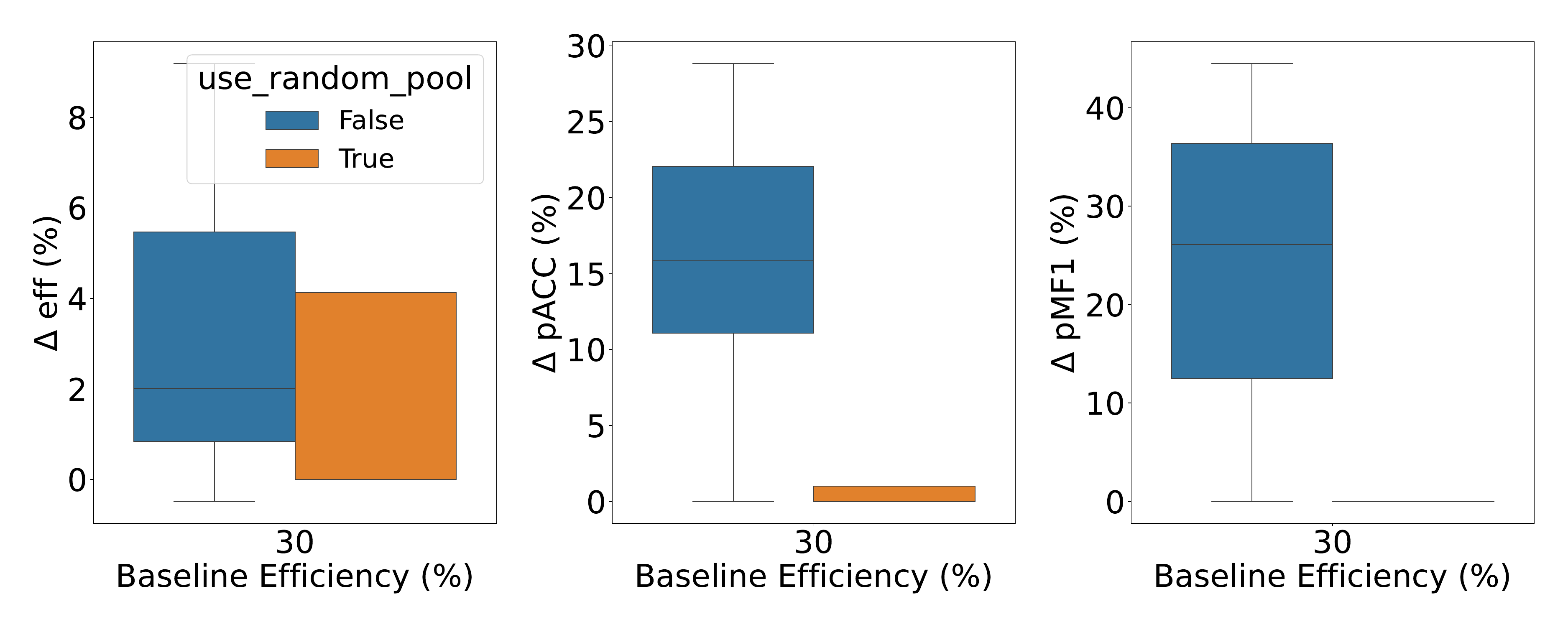}
        \caption{Results over categorical metrics for a setting where the design objective is set to maximize imputation performance.}
         \label{fig:ablation_random_pool_cat_mp}
    \end{subfigure}

    \begin{subfigure}[H]{\textwidth}
    \centering
    \includegraphics[width=\textwidth]{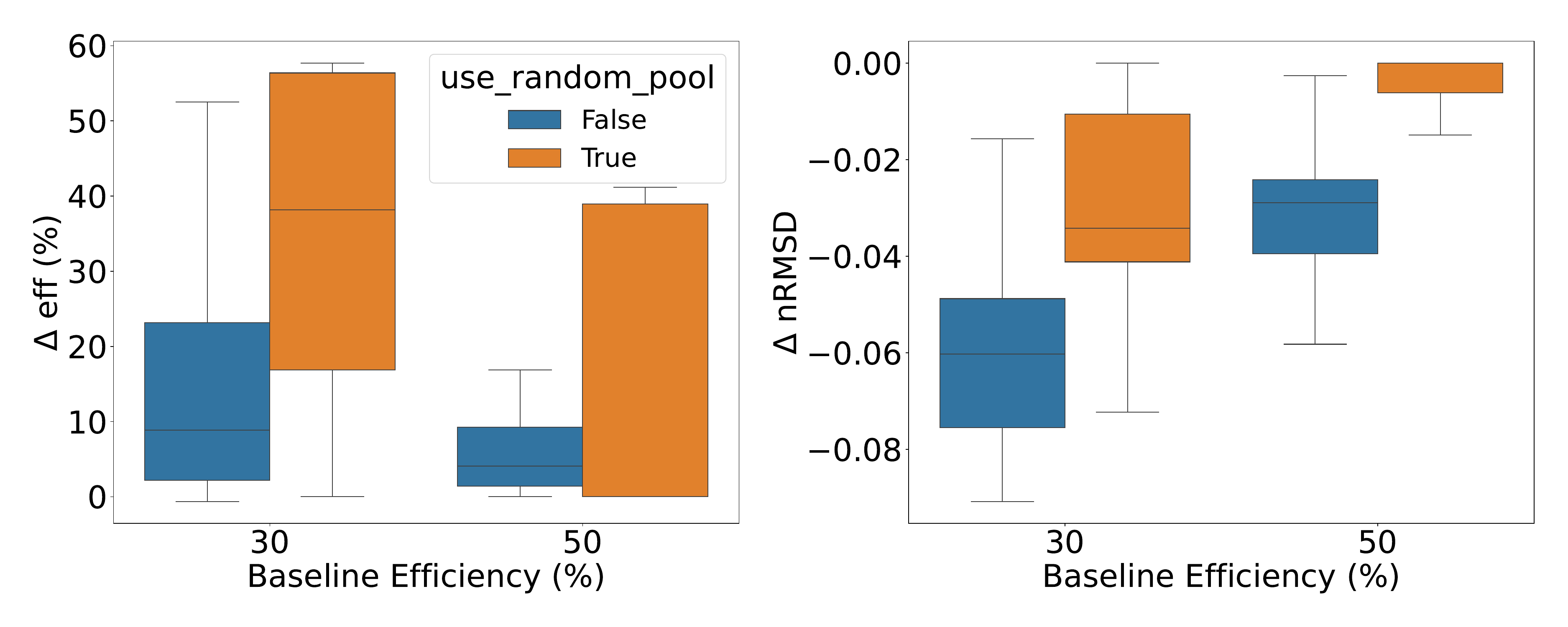}
    \caption{Results over continuous metrics for a setting where the design objective is set to maximize imputation performance.}
    \label{fig:ablation_random_pool_con_mp}
    \end{subfigure}

    \caption{Results comparing performance gains under METRIK against an ablated version that replaces the candidate pool of learned PMDs with random ones generated by the MF design. Gains are measured with respect to an MF-based PMD baseline. For simplicity, results are only shown for certain baseline efficiencies since the ablated method yields no performance gains at other baseline efficiencies.}
    \label{fig:ablation_random_pool}
    
\end{figure}

\begin{figure}[H]
\begin{subfigure}[H]{\textwidth}
    \centering
    \includegraphics[width=\textwidth]{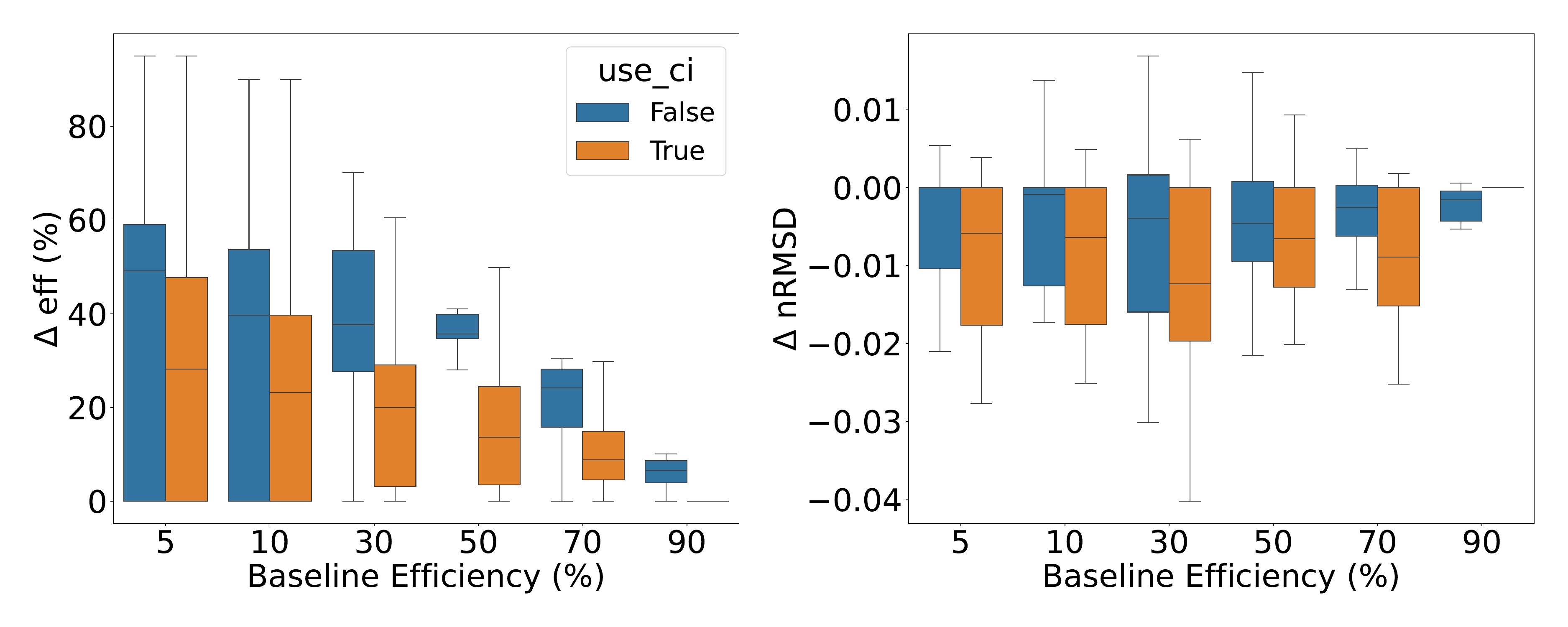}
    \caption{Results over continuous metrics for a setting where the design objective is set to maximize efficiency.}
    \label{fig:ci_ablation_con}
\end{subfigure}

\begin{subfigure}[H]{\textwidth}
    \centering
    \includegraphics[width=\textwidth]{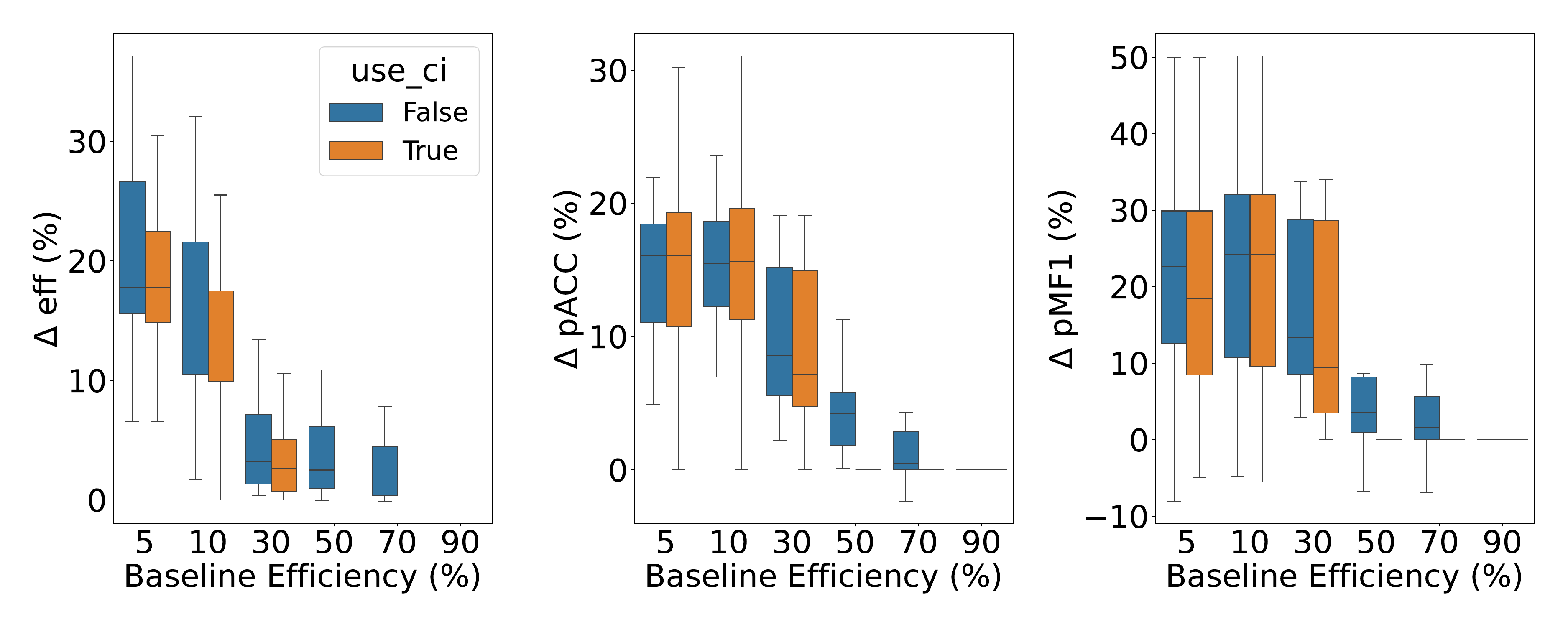}
    \caption{Results over categorical metrics for a setting where the design objective is set to maximize imputation performance.}
    \label{fig:ci_ablation_cat_mp}
\end{subfigure}

\begin{subfigure}[H]{\textwidth}
    \centering
    \includegraphics[width=\textwidth]{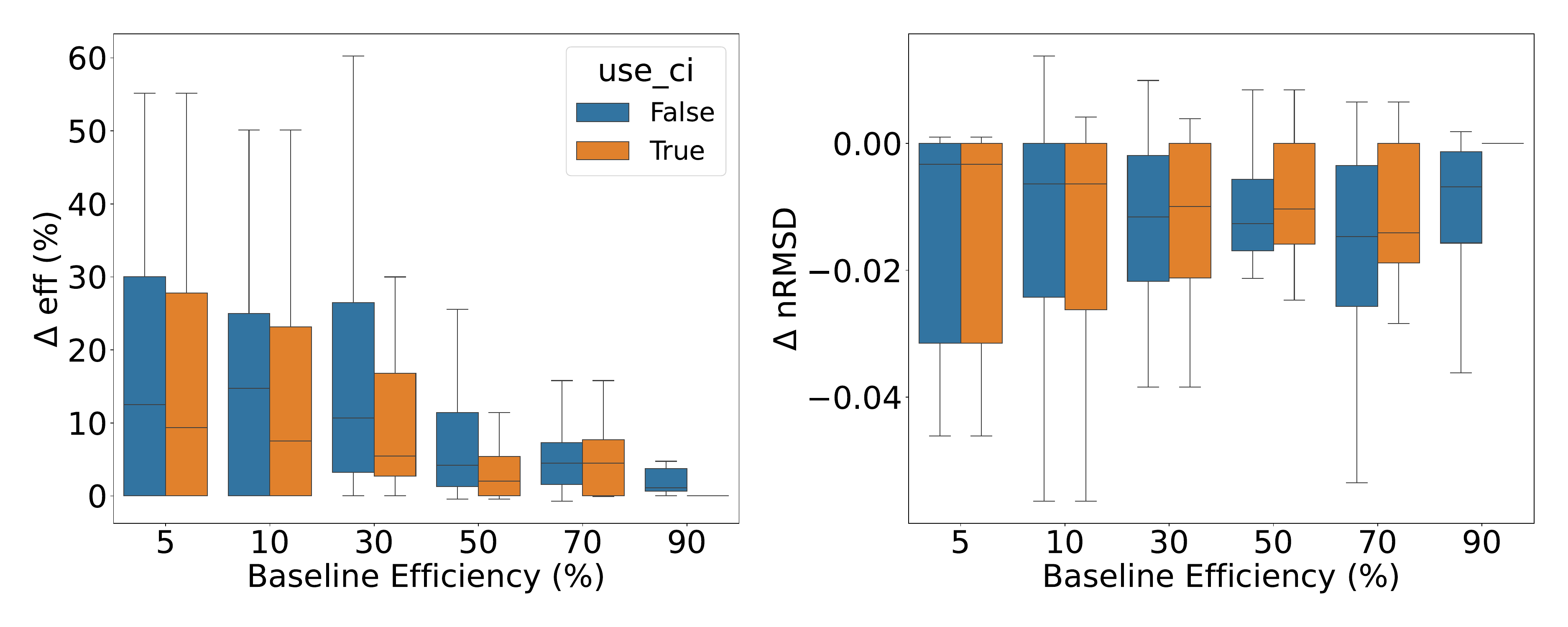}
    \caption{Results over continuous metrics where the design objective is set to maximize imputation performance.}
    \label{fig:ci_ablation_con_mp}
\end{subfigure}

\caption{Results comparing performance gains under METRIK against an ablated version that does not use confidence intervals for performance estimation during PMD selection. Gains are measured with respect to the RSD baseline.}
\label{fig:ci_ablation}

\end{figure}

\end{appendices}



\end{document}